\documentclass[pdflatex,sn-mathphys-num]{sn-jnl}

\usepackage{graphicx}%
\usepackage{multirow}%
\usepackage{amsmath,amssymb,amsfonts}%
\usepackage{amsthm}%
\usepackage{mathrsfs}%
\usepackage[title]{appendix}%
\usepackage{xcolor}%
\usepackage{textcomp}%
\usepackage{manyfoot}%
\usepackage{booktabs}%
\usepackage{algorithm}%
\usepackage{algorithmicx}%
\usepackage{algpseudocode}%
\usepackage{listings}%
\usepackage{tabularx}
\usepackage{booktabs}
\usepackage{array}
\usepackage{placeins}

\theoremstyle{thmstyleone}%
\theoremstyle{thmstyletwo}%

\theoremstyle{thmstylethree}%

\begin{document}

\title[Article Title]{Trustworthy Visual Predicates for Robust Manipulation Understanding under Degradation}


\author*[1]{\fnm{Fatemeh} \sur{Ziaeetabar}}\email{fziaeetabar@ut.ac.ir}

\affil[1]{
\orgdiv{Department of Computer Science},
\orgname{School of Mathematics, Statistics and Computer Science, College of Science, University of Tehran},
\orgaddress{
\city{Tehran},
\country{Iran}
}
}


\abstract{
Manipulation understanding requires reliable relational evidence, such as contact, support, containment, motion coupling, grasp, release, and active-hand involvement. Although these visual predicates are widely used in event-chain, graph-based, and neuro-symbolic models, their reliability under visual degradation is rarely analyzed directly. This paper introduces a predicate-level reliability framework for robust manipulation understanding under blur, occlusion, illumination change, low resolution, frame dropping, and detection noise. The framework defines a structured predicate vocabulary, confidence-aware predicate estimation, and reliability metrics for predicate preservation, degradation sensitivity, temporal consistency, confidence-weighted stability, and downstream impact. Experiments on controlled manipulation videos and public egocentric or bimanual datasets, including VISOR/EPIC-KITCHENS, H2O, and ARCTIC, show that predicate failures are structured rather than uniform. Static spatial predicates remain comparatively robust, whereas contact-sensitive, dynamic, and derived predicates such as grasp and release are more fragile. Under severe degradation, detection noise, occlusion, and frame dropping cause the strongest reliability losses. Downstream analysis shows that degraded predicates reduce manipulation-understanding accuracy from 0.89 to 0.58, while removing confidence weighting under moderate degradation reduces accuracy from 0.74 to 0.64. These results show that predicate reliability provides a diagnostic layer between visual perception and structured manipulation reasoning.
}

\keywords{visual predicate reliability, manipulation understanding, hand--object interaction, visual degradation, robustness analysis, egocentric video understanding, interpretable machine vision}

\maketitle

\section{Introduction}
\label{sec:introduction}

Manipulation understanding from visual observations is a central problem in machine vision. In applications such as assistive robotics, human--robot collaboration, industrial monitoring, egocentric video understanding, and interactive systems, a model must often interpret not only which objects are present, but also how hands, tools, containers, and surfaces interact over time \cite{Argall2009,damen2022epic,sener2022assembly101,grauman2024egoexo4d}. Recognizing a manipulation action therefore requires more than assigning a single label to a video clip. It requires inference of visually grounded relations such as approach, contact, coordinated movement, support, containment, and release.

We refer to such intermediate relations as \emph{visual predicates}. A visual predicate is a machine-vision-derived statement about a relation, state, or interaction in a video. Related relational representations have previously been explored in semantic event chains, hand--object interaction modeling, scene graphs, and visual relationship reasoning \cite{aksoy2011learning,ziaeetabar2017semantic,ziaeetabar2018recognition,krishna2017visual}. Examples include contact between a hand and an object, closeness between entities, coordinated movement of a hand and an object, containment, support, grasping, releasing, and active-hand involvement. These predicates are simpler than action labels, but they provide the perceptual evidence from which actions are interpreted. For example, ``putting a cup on a table'' may be described through contact, motion coupling, support, and release. If these predicates are unreliable, even a strong recognition model may fail for reasons that remain hidden when only final accuracy is reported.

Modern video understanding systems, including deep action recognition networks, graph neural networks, transformers, and vision foundation models, are commonly evaluated through final task performance \cite{Radford2021,liu2023grounding,kirillov2023segment}. This is useful for benchmarking, but it does not reveal why a manipulation action is misclassified. The error may originate from the classifier, temporal modeling, relational reasoning, or from unreliable visual predicates before reasoning begins. In manipulation understanding, this last source is critical because small errors in contact, containment, grasp, release, or motion relations can change the interpreted action.

This problem becomes more important under real-world visual degradation. Manipulation videos are often affected by motion blur, low illumination, overexposure, low resolution, compression artifacts, partial occlusion, frame dropping, missed detections, and unstable tracking. Robustness studies have shown that visual recognition systems can be sensitive to corruptions, noise, blur, and distribution shifts \cite{hendrycks2019benchmarking,geirhos2019imagenet,michaelis2019benchmarking}. However, these degradations affect manipulation predicates differently. Proximity may remain stable under moderate blur, whereas contact may fail when hand--object boundaries are unclear. Containment depends on segmentation and geometry, while motion predicates are sensitive to frame dropping and tracking errors. Robust manipulation understanding therefore requires predicate-level analysis: which predicates remain reliable, which fail, and how do their failures affect action interpretation?

\begin{figure}
\centering
\includegraphics[width=\textwidth]{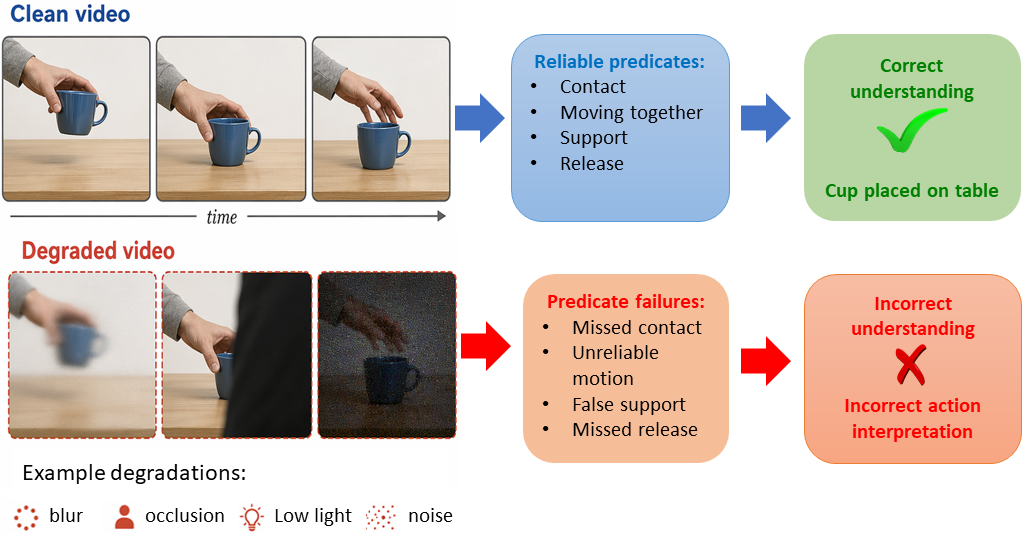}
\caption{Conceptual motivation of predicate-level reliability analysis. A clean manipulation video provides stable relational evidence, allowing predicates such as contact, moving together, support, and release to support the correct manipulation interpretation. Under visual degradation, the same 
action may produce unreliable predicate estimates, such as missed contact, unstable motion, false support, or missed release, which can propagate to incorrect or uncertain downstream interpretation.}
\label{fig:conceptual_motivation}
\end{figure}

This paper addresses this question by proposing a \emph{visual predicate reliability framework} for manipulation videos under real-world degradation. Instead of introducing another end-to-end action recognition model, the framework evaluates the stability of intermediate visual predicates and quantifies how degradation affects their reliability. The central claim is that manipulation understanding should be assessed not only by final action accuracy, but also by the trustworthiness of the predicates supporting the interpretation. Figure~\ref{fig:conceptual_motivation} illustrates this motivation by contrasting a clean-video pathway, where reliable predicates support correct understanding, with a degraded-video pathway, where predicate failures propagate to incorrect action interpretation.

\subsection{Motivation: Why Predicate Reliability Matters}
\label{subsec:motivation_predicate_reliability}

Manipulation actions are inherently relational. The same object may be touched, grasped, lifted, placed, inserted, removed, or released depending on contact, motion, spatial configuration, and temporal evolution. Thus, object recognition alone is insufficient; the system must also infer reliable relations among entities. This view is consistent with earlier relational and event-based formulations of manipulation, where action meaning is represented through changes in object relations rather than by appearance alone \cite{aksoy2011learning,ziaeetabar2017semantic,ziaeetabar2018recognition,ziaeetabar2020using,worgotter2020humans}.

Visual predicates provide the bridge between low-level perception and high-level action understanding. Perception modules provide detections, masks, keypoints, optical flow, trajectories, or feature embeddings, while action understanding requires labels, event descriptions, predictions, or explanations. Predicate-level information connects these levels by representing relations such as contact, support, containment, and coordinated movement.

This intermediate level is important for both interpretability and robustness. If a system predicts that an object was placed on a table, the prediction should be supported by evidence of contact, hand--object motion coupling, table support, and release. Similarly, if a system fails under blur, occlusion, or low light, predicate-level analysis can reveal whether the error was caused by contact instability, tracking failure, containment ambiguity, active-hand confusion, or temporal inconsistency. This diagnostic capability is useful for improving both perception modules and downstream reasoning systems.

Different downstream models may consume the same visual predicates, including end-to-end video models, graph-based systems, neuro-symbolic frameworks, and event-chain representations \cite{yan2018spatial,krishna2017visual,ziaeetabar2018recognition,ziaeetabar2024hierarchical,ziaeetabar2025adaptive,ziaeetabar2026neurosymbolic}. However, these models often assume that predicate inputs are reliable. This paper studies that assumption directly by evaluating predicate reliability rather than primarily focusing on the reasoning models that consume such predicates.

\subsection{Problem Statement}
\label{subsec:problem_statement}

The main research question is: \emph{How reliable are machine-vision-derived manipulation predicates under real-world visual degradation, and how do predicate failures affect downstream manipulation understanding?}

This question involves four aspects. First, some predicates may remain robust under degradation; for example, coarse proximity may still be estimated under moderate blur or low illumination. Second, fine-grained predicates such as contact, grasp, release, containment, and support may be fragile because they depend on accurate boundaries, stable localization, and temporal continuity. Third, degradation severity matters: mild blur or occlusion may have limited impact, while stronger degradation may remove the visual evidence needed for predicate estimation. Fourth, predicate errors may have different downstream consequences; for instance, a missed release event may be more harmful than a brief proximity error.

To study these issues, manipulation videos are processed to extract entities such as hands, objects, tools, containers, and surfaces. Predicate values are then estimated from visual evidence including spatial distance, mask overlap, bounding-box geometry, trajectory similarity, motion consistency, and visibility. Each predicate is associated with a confidence value, and the same videos are evaluated under clean and degraded conditions to measure reliability across degradation types and severity levels.

The considered degradation types are:

\begin{itemize}
\item \textbf{Blur:} motion blur and Gaussian blur reduce boundary sharpness.
\item \textbf{Occlusion:} hand--object and object--object occlusion hide interaction evidence.
\item \textbf{Illumination change:} low illumination and overexposure affect visibility and segmentation.
\item \textbf{Low resolution:} reduced spatial detail weakens evidence for contact, grasp, and containment.
\item \textbf{Temporal frame dropping:} missing frames disrupt motion continuity.
\item \textbf{Detection noise:} localization errors, missed detections, false detections, and tracking instability reduce predicate reliability.
\end{itemize}

The paper therefore does not primarily propose a new symbolic reasoning model, graph neural network, or foundation-model architecture. Instead, it proposes a reliability analysis layer for visual predicates. Downstream recognition or reasoning may be used as an evaluation case, but the main contribution is the analysis of predicate stability and failure under degradation.

\subsection{Main Contributions}
\label{subsec:contributions}

The main contribution of this paper is a predicate-level diagnostic framework for trustworthy manipulation understanding under visual degradation. Rather than treating robustness only as a change in final action-recognition accuracy, the proposed framework analyzes the reliability of the intermediate relational evidence on which manipulation understanding depends. The specific contributions are as follows.

\begin{enumerate}
    \item \textbf{Predicate-level reliability paradigm.}
    We formulate manipulation robustness as a predicate-level reliability problem. The framework treats visual predicates such as contact, support, containment, motion coupling, grasp, release, and active-hand involvement as the object of analysis, rather than assuming that such relations are reliable inputs to downstream reasoning.

    \item \textbf{Degradation-aware diagnostic framework.}
    We introduce a degradation-aware framework that compares predicate behavior between clean and degraded manipulation videos. The framework identifies stable, fragile, temporally inconsistent, and high-impact predicates across blur, occlusion, illumination change, low resolution, frame dropping, and detection noise.

    \item \textbf{Confidence-aware reliability metrics.}
    We define a unified reliability profile for each predicate, including Predicate Reliability Score ($\mathrm{PRS}$), Predicate Degradation Sensitivity ($\mathrm{PDS}$), Temporal Predicate Consistency ($\mathrm{TPC}$), Confidence-Weighted Predicate Stability ($\mathrm{CWS}$), and Downstream Impact Score ($\mathrm{DIS}$). Together, these metrics measure not only whether a predicate is preserved, but also whether its temporal structure, confidence, and downstream importance remain reliable.

    \item \textbf{Cross-dataset and downstream validation.}
    We evaluate the framework on controlled manipulation videos and public egocentric or bimanual manipulation datasets, including VISOR/EPIC-KITCHENS, H2O, and ARCTIC. The analysis shows that predicate failures are structured rather than uniform, and that failures of high-impact predicates such as grasp, release, contact, support, and moving together propagate to downstream manipulation-understanding performance.

    \item \textbf{Guidance for robust relational reasoning.}
    We show how predicate reliability can support graph-based, event-chain-based, and neuro-symbolic manipulation systems by indicating which relations should be trusted, down-weighted, assigned to $\mathrm{UNK}$, or refined before reasoning. The framework therefore provides a diagnostic layer between visual perception and structured manipulation understanding.
\end{enumerate}

The rest of the paper is organized as follows. Section~\ref{sec:relatedWorks} reviews related work. Section~\ref{sec:framework} presents the proposed visual predicate reliability framework. Section~\ref{sec:protocol} describes the experimental protocol. Section~\ref{sec:results} reports quantitative and qualitative results. Section~\ref{sec:discussion} discusses implications and limitations. Section~\ref{sec:conclusion} concludes the paper.

\section{Related Work}
\label{sec:relatedWorks}

The proposed predicate-level reliability framework lies at the intersection of video-based manipulation understanding, visual predicate extraction, robustness analysis, and relational reasoning. Accordingly, this section reviews prior work across these complementary directions, with particular attention to how existing methods represent intermediate relations and to the limited attention given to their reliability under degraded visual conditions. While previous research has substantially advanced action recognition, hand--object interaction analysis, egocentric video understanding, and symbolic or graph-based manipulation reasoning, most studies focus on final action labels, reconstructed poses, interaction classes, or model-level performance. In contrast, the present work focuses on the reliability of intermediate visual predicates that provide the evidence for robust manipulation understanding.

\subsection{Manipulation Understanding in Video}

Manipulation understanding differs from generic action recognition because action meaning often depends on fine-grained relations among hands, objects, tools, containers, and supporting surfaces. Actions such as grasping, placing, pouring, inserting, removing, and releasing may involve similar objects or global motion patterns, but differ in contact, motion coupling, containment, support, and object-state change. Therefore, manipulation understanding requires modeling not only what appears in a video, but also how entities interact over time.

General video action-recognition methods provide strong spatio-temporal representations, including two-stream networks, Inflated 3D ConvNets, Temporal Shift Modules, SlowFast networks, TimeSformer, and VideoMAE \cite{simonyan2014two,carreira2017quo,lin2019tsm,feichtenhofer2019slowfast,bertasius2021space,tong2022videomae}. These methods are effective for action classification, but they are usually evaluated through final recognition accuracy and provide limited diagnosis of the relational cues that support or undermine manipulation interpretation.

Egocentric and hand--object datasets have advanced manipulation-related video understanding. EPIC-KITCHENS-100 and Ego4D support large-scale egocentric activity analysis, while Assembly101, Ego-Exo4D, H2O, and ARCTIC provide multi-view, first-person, or bimanual manipulation data with varying degrees of hand, object, pose, and interaction information \cite{damen2022epic,grauman2022ego4d,sener2022assembly101,grauman2024egoexo4d,kwon2021h2o,fan2023arctic}. Recent hand--object interaction works further integrate hand pose, object pose, contact cues, and interaction recognition for bimanual or egocentric scenarios \cite{cho2023transformer,roh2023functional}. These studies demonstrate the importance of explicit hand--object modeling, but they mainly focus on action labels, poses, interaction classes, or reconstruction quality.

In contrast, the present work focuses on the reliability of intermediate visual predicates, such as contact, proximity, support, containment, grasp, release, and coordinated motion. These predicates provide the relational evidence required for robust manipulation understanding, yet their stability under visual degradation is rarely evaluated directly.

\subsection{Visual Predicate Extraction}

Visual predicate extraction converts low-level visual observations into intermediate relational statements about entities in a scene. In manipulation videos, such predicates may describe proximity, contact, motion coupling, support, containment, grasping, releasing, or active-hand involvement. They provide a bridge between perception outputs, such as detections, masks, keypoints, poses, trajectories, and optical flow, and higher-level action interpretation.

A broad basis for visual predicate extraction comes from visual relationship detection and scene-graph representation. Visual Genome introduced dense object, attribute, and relationship annotations for image-level scene graphs \cite{krishna2017visual}, while Action Genome extended this idea to spatio-temporal scene graphs in video \cite{ji2020action}. These works highlight the importance of relational visual representation, but their focus is mainly on relationship modeling or action recognition rather than on the reliability of extracted relations under degraded conditions.

In manipulation understanding, predicates are often derived from geometric, spatial, and temporal cues. Contact can be estimated from mask overlap or boundary distance, support and containment from object--surface or object--container geometry, and motion predicates from trajectory similarity or relative displacement. Datasets such as H2O, VISOR, ARCTIC, and ContactPose provide useful evidence for studying these cues through hand pose, object pose, segmentation masks, interaction labels, and contact annotations \cite{kwon2021h2o,darkhalil2022epic,fan2023arctic,brahmbhatt2020contactpose}. Segmentation and tracking methods, including promptable segmentation models such as Segment Anything, further provide the regions and correspondences required for predicate extraction \cite{kirillov2023segment}. However, errors in detection, segmentation, tracking, or pose estimation can directly affect fine-grained predicates such as contact, containment, support, grasp, and release.

Event-based manipulation representations are closely related to visual predicate extraction. Semantic Event Chains represent actions through changes in spatial relations among objects at decisive moments \cite{aksoy2011learning}, and Enriched Semantic Event Chains extend this idea using richer spatial and temporal relations for manipulation-action recognition and prediction \cite{ziaeetabar2017semantic,ziaeetabar2018prediction,ziaeetabar2018recognition,ziaeetabar2020using}. These approaches show that manipulation actions can be described through relational changes rather than appearance alone. However, they generally assume that the extracted relations are reliable. The present work studies this assumption directly by analyzing predicate reliability under blur, occlusion, illumination change, low resolution, frame dropping, and detection noise.

\subsection{Robustness under Visual Degradation}

Robustness under visual degradation has become an important topic in machine vision because models trained and evaluated on clean data often fail under realistic acquisition conditions. Common corruptions such as noise, blur, illumination change, compression artifacts, resolution loss, and occlusion can substantially reduce recognition performance. ImageNet-C and ImageNet-P introduced standardized benchmarks for evaluating image-classification robustness under common corruptions and perturbations \cite{hendrycks2019benchmarking}. Related studies further showed that deep networks may rely strongly on texture cues, which can limit their robustness under distribution shifts and degraded visual conditions \cite{geirhos2019imagenet}.

Robustness has also been studied for object detection and video understanding. For object detection, corruption benchmarks such as PASCAL-C, COCO-C, and Cityscapes-C demonstrated that detectors can suffer significant performance loss under noise, blur, weather, and digital artifacts \cite{michaelis2019benchmarking}. In video recognition, robustness is more complex because degradations may affect both spatial appearance and temporal continuity. Video-specific studies have therefore considered corruptions such as motion blur, frame-rate changes, bit errors, packet loss, compression, and temporal discontinuities \cite{hirata2021making,zeng2024benchmarking}. These works show that robustness should be evaluated not only on clean videos, but also under realistic spatial and temporal degradation.

For manipulation understanding, degradation is especially critical because many actions depend on fine-grained visual evidence. Blur or low resolution may obscure hand--object boundaries, occlusion may hide contact or release events, illumination changes may affect segmentation, and frame dropping may disrupt motion continuity. As a result, different predicates are expected to fail in different ways. For example, proximity may remain relatively stable under moderate blur, while contact, grasp, containment, or release may be more fragile because they require accurate localization, visibility, and temporal consistency.

Most robustness studies evaluate performance at the level of image classes, object detections, or final video action labels. However, manipulation understanding requires reliable intermediate relations before high-level reasoning can succeed. The present work therefore extends robustness analysis to the predicate level by studying how visual degradation affects relational cues such as contact, support, containment, grasp, release, active-hand involvement, and coordinated motion.

\subsection{Relational and Symbolic Representations}

Relational and symbolic representations provide an interpretable way to model manipulation actions by describing how objects, hands, tools, and surfaces are related over time. Instead of representing an action only as a global video label, these approaches encode intermediate structures such as spatial relations, object states, temporal transitions, interaction graphs, or symbolic predicates. This is particularly relevant for manipulation understanding, where the meaning of an action often depends on changes in relations such as contact, support, containment, grasp, release, and coordinated motion.

Semantic Event Chains (SECs) are among the most influential relational representations for manipulation analysis. They describe actions through changes in spatial relations among objects at decisive temporal moments \cite{aksoy2011learning}. Enriched Semantic Event Chains (eSECs) extend this idea by incorporating richer spatial and temporal information for manipulation-action recognition and prediction \cite{ziaeetabar2017semantic,ziaeetabar2018prediction,ziaeetabar2018recognition,ziaeetabar2020using}. These methods show that manipulation can be understood as a sequence of relational changes rather than as appearance-based classification alone. However, classical SEC/eSEC formulations usually assume that the extracted visual relations are sufficiently reliable.

Graph-based representations offer another way to encode relational structure. Spatio-temporal graph models have been widely used for action recognition by representing entities, body joints, or scene elements as nodes and their relations as edges \cite{yan2018spatial}. In manipulation analysis, hierarchical graph-based models have been used to represent object-level, single-hand, and bimanual action information for recognition and description generation \cite{ziaeetabar2024hierarchical}. More recent multimodal graph reasoning approaches further combine relational structures with foundation-model representations and language-based contextual information for fine-grained action recognition \cite{ziaeetabar2025adaptive}. These works demonstrate the value of structured relational modeling, but they mainly focus on recognition, reasoning, or description generation.

Neuro-symbolic approaches aim to combine the interpretability of symbolic representations with the flexibility of learning-based perception. Recent work on neuro-symbolic manipulation understanding with enriched Semantic Event Chains proposes to use eSEC-based symbolic states together with confidence-aware predicates, functional roles, affordance priors, and explanation cues for manipulation reasoning \cite{ziaeetabar2026neurosymbolic}. This direction is closely related to the present study because it highlights the importance of predicates as internal representations for reasoning. However, the focus of the present paper is different: rather than proposing a new symbolic or neuro-symbolic reasoning architecture, it studies whether the visual predicates that such systems rely on remain reliable under real-world degradation.

Overall, relational, graph-based, and neuro-symbolic methods demonstrate that intermediate predicates can support interpretable manipulation understanding. Nevertheless, most of these approaches treat predicates as inputs to reasoning systems and do not systematically evaluate their robustness. The proposed work addresses this gap by analyzing predicate reliability before these predicates are used by symbolic, graph-based, or multimodal reasoning modules. Table~\ref{tab:positioning} summarizes the position of the proposed work across the four related directions.

\begin{table}[!htbp]
\centering
\caption{Positioning of the proposed work with respect to major related directions.}
\label{tab:positioning}
\small
\begin{tabularx}{\linewidth}{>{\raggedright\arraybackslash}p{2.8cm}
                                >{\raggedright\arraybackslash}X
                                >{\raggedright\arraybackslash}X
                                >{\raggedright\arraybackslash}X}
\toprule
\textbf{Direction} & 
\textbf{Main Focus} & 
\textbf{Typical Limitation} & 
\textbf{Proposed Focus} \\
\midrule

Manipulation understanding 
& Action, hand--object, egocentric, and bimanual recognition 
& Mainly final labels, poses, or segments 
& Reliability of relational evidence \\

Visual predicate extraction 
& Spatial, contact, motion, and scene-graph relations 
& Predicates used but rarely evaluated directly 
& Predicate-level reliability analysis \\

Robust vision 
& Performance under visual or temporal degradation 
& Mostly class-, detection-, or action-level robustness 
& Degradation-aware predicate analysis \\

Relational/symbolic models 
& Event chains, graphs, and neuro-symbolic reasoning 
& Often assumes reliable predicate inputs 
& Trustworthiness of predicate inputs \\

\bottomrule
\end{tabularx}
\end{table}

\section{Proposed Visual Predicate Reliability Framework}
\label{sec:framework}

This section presents the proposed framework for analyzing the reliability of visual predicates in manipulation videos under degraded visual conditions. The framework builds on the relational view of manipulation used in SEC/eSEC-based representations, where actions are described through changes in contact, static spatial relations, and dynamic spatial relations among interacting entities \cite{ziaeetabar2017semantic,ziaeetabar2018prediction,ziaeetabar2018recognition,ziaeetabar2020using}. However, unlike prior works that use such relations mainly for recognition, prediction, or reasoning, the present framework treats the predicates themselves as the object of analysis and evaluates their stability, confidence, and downstream impact under visual degradation.

Given a manipulation video, the framework extracts task-relevant entities, computes pairwise or group-wise predicates, assigns confidence scores, and compares predicate behavior between clean and degraded conditions. This enables the analysis of which predicates remain reliable, which become fragile, and which failures most affect manipulation understanding. The following subsections define the framework notation, entity representation, predicate vocabulary, confidence estimation, reliability metrics, and downstream use.

\subsection{Framework Overview and Notation}
\label{subsec:framework_overview}

\begin{figure*}[t]
\centering
\includegraphics[width=\linewidth]{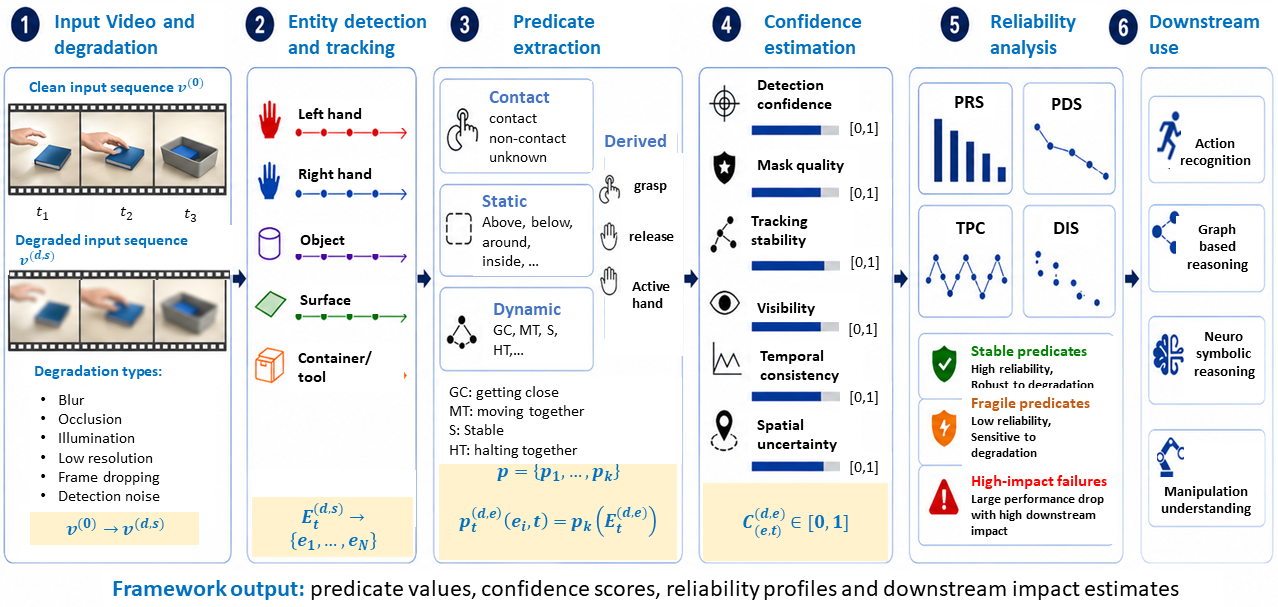}
\caption{Overview of the proposed visual predicate reliability framework. A clean manipulation video and its degraded variants are processed through entity extraction, predicate estimation, confidence scoring, and reliability analysis. The resulting reliability profile supports downstream manipulation understanding by identifying stable, fragile, and high-impact predicates.}
\label{fig:framework_overview}
\end{figure*}

Figure~\ref{fig:framework_overview} illustrates the overall pipeline of the proposed visual predicate reliability framework. Given a manipulation video, the framework generates degraded variants of the same sequence, extracts task-relevant entities, computes SEC/eSEC-inspired visual predicates, estimates predicate confidence, and evaluates predicate reliability across degradation types and severity levels. The output is a predicate-level reliability profile that identifies stable predicates, fragile predicates, and predicate failures that may affect downstream manipulation understanding.

Let
\begin{equation}
V^{0}=\{I^{0}_{t}\}_{t=1}^{T}
\end{equation}
denote a clean manipulation video with $T$ frames, where $I^{0}_{t}$ is the clean frame at time $t$. For each degradation type $d \in \mathcal{D}$ and severity level $s \in \mathcal{S}$, the corresponding degraded video is denoted as
\begin{equation}
V^{d,s}=\{I^{d,s}_{t}\}_{t=1}^{T}.
\end{equation}
The degradation set $\mathcal{D}$ may include blur, occlusion, illumination change, low resolution, temporal frame dropping, and detection noise. Comparing predicate estimates from $V^{0}$ and $V^{d,s}$ allows the framework to quantify how visual degradation affects the relational evidence used for manipulation understanding.

At each frame $t$, a perception module extracts a set of visual entities:
\begin{equation}
E_t^{d,s} = \{e^{d,s,t}_1,e^{d,s,t}_2,\ldots,e^{d,s,t}_{N_t}\},
\end{equation}
where each entity may correspond to a hand, manipulated object, tool, container, supporting surface, or target object. Each entity is represented by visual evidence such as a class label, bounding box, segmentation mask, track identity, detection confidence, visibility estimate, and trajectory information. Predicates are then computed over single entities, entity pairs, or small entity groups.

Following the SEC/eSEC view of manipulation, the base predicate space is organized into contact, static spatial, and dynamic spatial relations. We denote these relation sets as
\begin{equation}
\mathcal{R}^{C}=\{\mathrm{T},\,\mathrm{N},\,\mathrm{UNK}\},
\end{equation}
\begin{equation}
\mathcal{R}^{S}=\{\mathrm{Ab},\,\mathrm{Be},\,\mathrm{Ar},\,\mathrm{In},\,\mathrm{Sup},\,\mathrm{UNK}\},
\end{equation}
and
\begin{equation}
\mathcal{R}^{D}=\{\mathrm{GC},\,\mathrm{MA},\,\mathrm{MT},\,\mathrm{FMT},\,\mathrm{HT},\mathrm{S},\,\mathrm{UNK}\}.
\end{equation}
Here, $\mathrm{T}$ and $\mathrm{N}$ denote contact and non-contact; $\mathrm{Ab}$, $\mathrm{Be}$, $\mathrm{Ar}$, $\mathrm{In}$, and $\mathrm{Sup}$ denote above, below, around, inside, and support/on-surface; and $\mathrm{GC}$, $\mathrm{MA}$, $\mathrm{MT}$, $\mathrm{FMT}$, $\mathrm{HT}$, and $\mathrm{S}$ denote getting close, moving apart, moving together, fixed-moving together, halting together, and stable. The $\mathrm{UNK}$ state is used when the visual evidence is insufficient, for example due to occlusion, missed detection, unreliable segmentation, or ambiguous motion.

Let $\mathcal{P}=\{p_1,p_2,\ldots,p_K\}$ denote the full predicate vocabulary used by the framework. This vocabulary includes the SEC/eSEC-inspired base relations above, together with derived manipulation predicates such as grasp, release, active-hand involvement, containment events, and support events. A predicate estimate for predicate $p_k$ at time $t$ under degradation condition $(d,s)$ is written as
\begin{equation}
\hat{y}^{d,s}_{k,t}=p_k(E_t^{d,s}),
\end{equation}
where $\hat{y}^{d,s}_{k,t}$ may be binary, categorical, or continuous depending on the predicate type. For example, contact/non-contact may be represented as a categorical or binary predicate, above/below/around/inside as static spatial predicates, and getting-close or moving-together as dynamic predicates. Derived predicates such as grasp or release are computed from combinations of contact, distance, motion coupling, and temporal change.

Motivated by recent confidence-aware neuro-symbolic extensions of eSEC representations \cite{ziaeetabar2026neurosymbolic}, each predicate estimate is associated with a confidence score
\begin{equation}
c^{d,s}_{k,t} \in [0,1],
\end{equation}
which reflects the reliability of the underlying visual evidence used to compute that predicate. The complete framework consists of six main stages: video input and degradation generation, entity detection and tracking, predicate extraction, confidence estimation, reliability analysis, and downstream use. These stages provide the technical basis for the entity representation, predicate vocabulary, confidence model, and reliability metrics defined in the following subsections.

\subsection{Visual Entity Extraction}
\label{subsec:visual_entity_extraction}

The first computational stage is the extraction of task-relevant visual entities from each frame. Since the proposed analysis is predicate-centered, entity extraction is not treated as the final objective, but as the perceptual basis for computing relational predicates. At each frame $t$ of a clean or degraded video $V^{d,s}$, the perception module produces the entity set $E_t^{d,s}$ defined in Section~\ref{subsec:framework_overview}. These entities correspond to physical participants that may enter into contact, spatial, or motion relations during a manipulation episode.

The entity vocabulary includes the left hand, right hand, manipulated object, tool, container, supporting surface, and target object. Not all entity types are present in every sequence, and distractor objects are excluded from predicate computation unless they participate in task-relevant relations. Entity extraction is performed using a perception front-end that provides recognition, localization, segmentation, and temporal tracking. In the implementation, this front-end may use vision foundation models for open-vocabulary recognition, semantic grounding, or promptable segmentation, together with task-specific hand detection, pose estimation, instance segmentation, or tracking modules \cite{Radford2021,liu2023grounding,kirillov2023segment}. The framework is independent of a particular detector or segmenter; it only requires entity-level evidence from which predicates and confidence scores can be computed.

Each detected entity $e_i^{d,s,t} \in E_t^{d,s}$ is represented as
\begin{equation}
e_i^{d,s,t} =
\left(
\ell_i^{d,s,t},
b_i^{d,s,t},
m_i^{d,s,t},
\tau_i^{d,s,t},
q_i^{d,s,t},
v_i^{d,s,t},
\mathbf{x}_i^{d,s,t}
\right),
\end{equation}
where $\ell_i^{d,s,t}$ denotes the entity label, $b_i^{d,s,t}$ the bounding box, $m_i^{d,s,t}$ the segmentation mask, $\tau_i^{d,s,t}$ the track identity, $q_i^{d,s,t} \in [0,1]$ the detection or recognition confidence, $v_i^{d,s,t} \in [0,1]$ the visibility estimate, and $\mathbf{x}_i^{d,s,t}$ a geometric state such as the object center, mask centroid, keypoint location, or trajectory descriptor. The label, box, and mask provide the semantic and geometric evidence required for contact and static spatial predicates, while the track identity and geometric state support temporal predicates.

Entity trajectories are obtained by linking geometric states across frames using the track identity:
\begin{equation}
\Gamma_i^{d,s} =
\left\{
\mathbf{x}_i^{d,s,t}
\right\}_{t=1}^{T}.
\end{equation}
These trajectories are used to estimate dynamic relations such as $\mathrm{GC}$, $\mathrm{MA}$, $\mathrm{MT}$, $\mathrm{FMT}$, $\mathrm{HT}$, and $\mathrm{S}$.

Entity extraction under degradation is expected to be imperfect. Blur may reduce boundary sharpness, occlusion may lower visibility, illumination changes may affect recognition and segmentation, and frame dropping may interrupt tracking. The framework therefore retains confidence, visibility, mask, and tracking information instead of converting detections directly into hard symbolic relations. These quantities are later used for predicate confidence estimation and reliability scoring.

The output of this stage is a structured entity representation for predicate computation: contact predicates use masks, overlap, or boundary distances; static spatial predicates use boxes, masks, or geometric states; and dynamic predicates use trajectories and relative motion. This representation provides the perceptual foundation for the predicate vocabulary and extraction rules defined next.

\subsection{Predicate Vocabulary and Extraction Rules}
\label{subsec:predicate_vocabulary}

Given the entity representation defined in Section~\ref{subsec:visual_entity_extraction}, the next stage is to convert entity-level visual evidence into relational predicates. The predicate vocabulary is designed to remain consistent with the SEC/eSEC view of manipulation, where actions are represented through changes in contact, static spatial relations, and dynamic spatial relations. At the same time, the framework also includes derived manipulation predicates, such as grasp and release, which are useful for downstream action interpretation.

For a pair of entities $(e_i^{d,s,t},e_j^{d,s,t}) \in E_t^{d,s}$, the framework estimates three groups of base predicates:
\begin{equation}
\hat{r}^{C,d,s}_{ij,t} \in \mathcal{R}^{C}, \qquad
\hat{r}^{S,d,s}_{ij,t} \in \mathcal{R}^{S}, \qquad
\hat{r}^{D,d,s}_{ij,t} \in \mathcal{R}^{D},
\end{equation}
where $\mathcal{R}^{C}$, $\mathcal{R}^{S}$, and $\mathcal{R}^{D}$ were defined in Section~\ref{subsec:framework_overview}. The contact relation $\hat{r}^{C,d,s}_{ij,t}$ describes whether two entities are physically interacting, the static spatial relation $\hat{r}^{S,d,s}_{ij,t}$ describes their spatial configuration at time $t$, and the dynamic spatial relation $\hat{r}^{D,d,s}_{ij,t}$ describes their relative motion over a short temporal window.

Contact predicates are estimated from geometric evidence such as mask overlap, boundary distance, depth proximity, or contact annotations when available. For example, a contact relation between two entities may be estimated as
\begin{equation}
\hat{r}^{C,d,s}_{ij,t} =
\begin{cases}
\mathrm{T}, & \Delta(m_i^{d,s,t},m_j^{d,s,t}) \leq \theta_C,\\
\mathrm{N}, & \Delta(m_i^{d,s,t},m_j^{d,s,t}) > \theta_C,\\
\mathrm{UNK}, & \text{if visual evidence is insufficient},
\end{cases}
\end{equation}
where $\Delta(\cdot,\cdot)$ denotes a boundary-distance, mask-overlap, or depth-based contact measure, and $\theta_C$ is a contact threshold. The unknown state is assigned when one or both entities are poorly localized, strongly occluded, or missing.

Static spatial predicates are computed from bounding boxes, masks, centroids, or 3D object states. Relations such as $\mathrm{Ab}$, $\mathrm{Be}$, $\mathrm{Ar}$, and $\mathrm{In}$ are determined from the relative position and containment geometry of the entities. For example, an object may be classified as $\mathrm{Ab}$ if its centroid or lower boundary lies above another entity, $\mathrm{In}$ if its mask or bounding region is contained within a container region, and $\mathrm{Ar}$ when it is laterally or spatially near another entity without satisfying a more specific containment or support relation. The support/on-surface relation $\mathrm{Sup}$ is used when an object is spatially above a surface while maintaining stable contact or near-contact with it.

Dynamic spatial predicates are computed from trajectories over a short temporal window. Let $\Gamma_i^{d,s}$ and $\Gamma_j^{d,s}$ denote the trajectories of two entities. The framework estimates relative motion using changes in distance, velocity similarity, and motion consistency. For example, if the distance between two entities decreases over time, the relation may be assigned as $\mathrm{GC}$; if it increases, it may be assigned as $\mathrm{MA}$; and if both entities move with similar trajectories while maintaining contact or proximity, the relation may be assigned as $\mathrm{MT}$. Fixed-moving-together $\mathrm{FMT}$ describes the case in which one entity remains approximately fixed while the other moves in contact with or along it. Halting-together $\mathrm{HT}$ indicates that two previously moving entities become jointly stationary, while $\mathrm{S}$ denotes stable relative configuration.

In addition to these base relations, the framework defines derived manipulation predicates that combine contact, spatial configuration, and temporal change. These predicates are not independent low-level detections; rather, they are inferred from the base predicate streams. For example, a grasp predicate can be estimated when a hand is close to or in contact with an object and the hand--object pair subsequently exhibits coordinated motion:
\begin{equation}
\hat{y}^{d,s}_{\mathrm{grasp},t}=1
\quad \text{if} \quad
\hat{r}^{C,d,s}_{h o,t}=\mathrm{T}
\ \land\
\hat{r}^{D,d,s}_{h o,t} \in \{\mathrm{MT},\mathrm{FMT}\},
\end{equation}
where $h$ denotes a hand entity and $o$ a manipulated object. Similarly, release can be detected when a previously contacted or grasped object changes to non-contact and the hand and object begin moving apart or become dynamically decoupled:
\begin{equation}
\hat{y}^{d,s}_{\mathrm{release},t}=1
\quad \text{if} \quad
\hat{r}^{C,d,s}_{h o,t-1}=\mathrm{T}
\ \land\
\hat{r}^{C,d,s}_{h o,t}=\mathrm{N}.
\end{equation}
Active-hand involvement is estimated by comparing the contact, proximity, and motion-coupling evidence of the left and right hands with respect to the manipulated object. The active hand is the hand with the strongest and most reliable relational evidence during the current temporal window.

\begin{figure*}[!t]
\centering
\includegraphics[width=\textwidth]{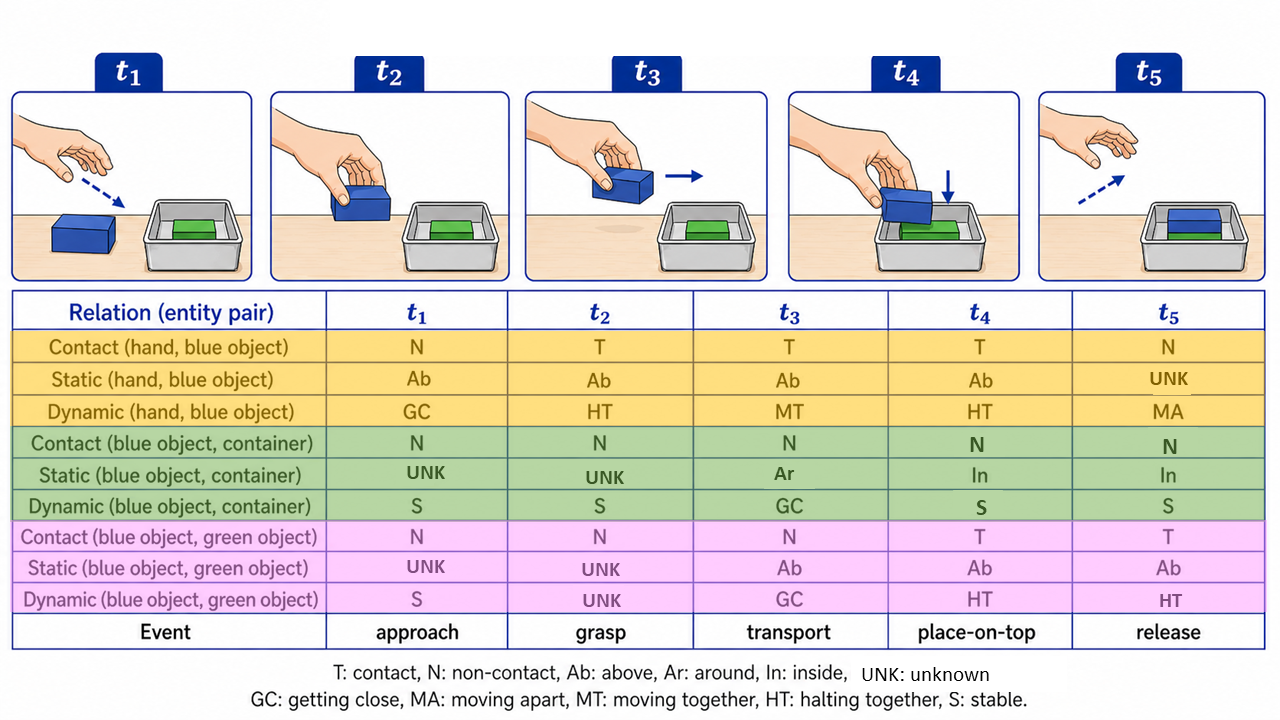}
\caption{Illustrative example of predicate extraction over time. A blue object is manipulated by the hand and placed on top of a green object that remains inside the container throughout the sequence. For each time step, the framework assigns contact, static spatial, and dynamic spatial relations for relevant entity pairs, including (hand, blue object), (blue object, container), and (blue object, green object). The example shows how predicate streams evolve during the action and how higher-level manipulation events, such as approach, transport, place-on-top, and release, can be inferred from them.}
\label{fig:predicate_extraction_example}
\end{figure*}

Figure~\ref{fig:predicate_extraction_example} illustrates a representative predicate-extraction example. A blue object is manipulated by the hand and placed on top of a green object that remains inside the container throughout the sequence. For each time step, the framework assigns contact, static spatial, and dynamic spatial relations for relevant entity pairs, such as (hand, blue object), (blue object, container), and (blue object, green object). This example highlights how low-level relational predicates evolve over time and how derived events such as grasp, place-on-top, and release can be inferred from these predicate streams.

Table~\ref{tab:predicate_vocabulary} summarizes the main predicate groups, their visual evidence, and typical failure modes under visual degradation. These extraction rules are intentionally modular: they may be implemented using hand-crafted geometric thresholds, learned predicate classifiers, or hybrid rules that combine learned perception with symbolic relation definitions. The reliability framework does not require a specific implementation of the predicate extractor; it requires that each predicate estimate $\hat{y}^{d,s}_{k,t}$ and its supporting evidence be available for confidence estimation and reliability analysis.

\begin{table}[!b]
\centering
\caption{Predicate vocabulary and visual evidence used in the proposed framework.}
\label{tab:predicate_vocabulary}
\small
\begin{tabular}{p{2.7cm} p{4.0cm} p{4.2cm}}
\hline
\textbf{Predicate group} & \textbf{Visual evidence} & \textbf{Typical failure mode} \\
\hline
Contact / non-contact 
& Mask overlap, boundary distance, depth proximity 
& Occlusion, blur, poor segmentation \\

Static spatial relations 
& Bounding boxes, masks, centroids, object geometry 
& Low resolution, localization error, partial visibility \\

Dynamic spatial relations 
& Trajectory similarity, distance change, motion consistency 
& Frame dropping, tracking instability, motion blur \\

Grasp 
& Contact, proximity, hand--object motion coupling 
& Boundary ambiguity, missed contact, hand occlusion \\

Release 
& Transition from contact to non-contact, motion decoupling 
& Missed frames, unstable tracking, temporary occlusion \\

Active hand 
& Relative contact, proximity, and motion evidence of both hands 
& Hand confusion, crossed hands, asymmetric occlusion \\
\hline
\end{tabular}
\end{table}

\subsection{Predicate Confidence Estimation}
\label{subsec:predicate_confidence}

Predicate extraction should not produce only a discrete relational label. Under degraded visual conditions, the same predicate value may be supported by strong, weak, or ambiguous evidence. Therefore, each predicate estimate $\hat{y}^{d,s}_{k,t}$ is associated with a confidence score $c^{d,s}_{k,t}\in[0,1]$, as introduced in Section~\ref{subsec:framework_overview}. This score represents the trustworthiness of the visual evidence used to estimate predicate $p_k$ at time $t$ under degradation condition $(d,s)$.

The confidence score is computed from an evidence-quality vector
\begin{equation}
\mathbf{z}^{d,s}_{k,t} =
\left[
q^{d,s}_{\mathrm{det},k,t},
q^{d,s}_{\mathrm{seg},k,t},
q^{d,s}_{\mathrm{trk},k,t},
q^{d,s}_{\mathrm{vis},k,t},
q^{d,s}_{\mathrm{sp},k,t},
q^{d,s}_{\mathrm{tmp},k,t}
\right],
\end{equation}
where the components denote detection or recognition confidence, segmentation or mask quality, tracking stability, visibility, spatial certainty, and temporal consistency, respectively. Each component is normalized to $[0,1]$, with larger values indicating stronger evidence.

A general confidence estimator is written as
\begin{equation}
c^{d,s}_{k,t}=g_k\left(\mathbf{z}^{d,s}_{k,t}\right),
\end{equation}
where $g_k(\cdot)$ is a predicate-specific confidence function. In the rule-based implementation used here, this function is defined as a weighted combination:
\begin{equation}
c^{d,s}_{k,t}
=
\sum_{r=1}^{R} w_{k,r} z^{d,s}_{k,t,r},
\qquad
\sum_{r=1}^{R} w_{k,r}=1,
\end{equation}
where $z^{d,s}_{k,t,r}$ is the $r$-th evidence-quality component and $w_{k,r}$ is its predicate-specific weight. The weights are selected according to the evidence requirements of each predicate group. For example, contact and grasp depend strongly on segmentation, visibility, and spatial certainty, whereas dynamic predicates such as $\mathrm{MT}$ and $\mathrm{MA}$ depend more strongly on tracking stability and temporal consistency.

The evidence components are computed from the entity representation defined in Section~\ref{subsec:visual_entity_extraction}. For a pairwise predicate over entities $e_i$ and $e_j$, detection confidence and visibility can be defined as
\begin{equation}
q^{d,s}_{\mathrm{det},k,t}
=
\min
\left(
q_i^{d,s,t},
q_j^{d,s,t}
\right),
\end{equation}
and
\begin{equation}
q^{d,s}_{\mathrm{vis},k,t}
=
\min
\left(
v_i^{d,s,t},
v_j^{d,s,t}
\right),
\end{equation}
so that predicate confidence decreases when either entity is weakly detected or poorly visible.

Segmentation quality $q^{d,s}_{\mathrm{seg},k,t}$ reflects the reliability of masks or object boundaries and is especially important for contact, containment, and support predicates. Tracking stability $q^{d,s}_{\mathrm{trk},k,t}$ measures whether the involved entities preserve consistent identities over time, using cues such as track continuity, identity-switch frequency, or trajectory smoothness.

Spatial certainty $q^{d,s}_{\mathrm{sp},k,t}$ measures how far the visual evidence is from a predicate decision boundary. For a distance- or overlap-based predicate, it can be defined as
\begin{equation}
q^{d,s}_{\mathrm{sp},k,t}
=
\min
\left(
1,
\frac{\left|\Delta^{d,s}_{ij,t}-\theta_k\right|}{\eta_k}
\right),
\end{equation}
where $\Delta^{d,s}_{ij,t}$ is the relevant distance or overlap measure, $\theta_k$ is the predicate threshold, and $\eta_k$ controls the uncertainty margin around the threshold.

Temporal consistency $q^{d,s}_{\mathrm{tmp},k,t}$ measures whether the predicate estimate is stable within a local temporal window. For binary or categorical predicates, it is computed as
\begin{equation}
q^{d,s}_{\mathrm{tmp},k,t}
=
\frac{1}{|\mathcal{W}_t|}
\sum_{\tau \in \mathcal{W}_t}
\mathbb{I}
\left(
\hat{y}^{d,s}_{k,\tau}
=
\hat{y}^{d,s}_{k,t}
\right),
\end{equation}
where $\mathcal{W}_t$ is a local temporal window around frame $t$ and $\mathbb{I}(\cdot)$ is the indicator function.

The confidence score is not itself a reliability metric. It describes the quality of evidence supporting a predicate estimate at a specific time and degradation condition. The reliability metrics introduced next use predicate values and confidence scores across clean and degraded videos to quantify stability, degradation sensitivity, temporal consistency, and downstream impact.

\subsection{Predicate Reliability Metrics}
\label{subsec:predicate_reliability_metrics}

The previous subsections define how visual predicates are extracted and how each predicate estimate is assigned a confidence score. We now define the reliability metrics used to quantify how predicates behave under visual degradation. The purpose of these metrics is to evaluate predicates not only as isolated frame-level outputs, but also as temporally evolving relational evidence that supports downstream manipulation understanding.

For each predicate $p_k \in \mathcal{P}$, the framework compares the predicate sequence extracted from the clean video $V^0$ with the corresponding sequence extracted from a degraded video $V^{d,s}$. We denote the clean predicate estimate at time $t$ as $\hat{y}^{0}_{k,t}$ and the degraded predicate estimate as $\hat{y}^{d,s}_{k,t}$. When available, ground-truth predicate labels can be used instead of clean-video estimates. However, in many practical settings, clean-video predicates provide the reference against which degradation-induced changes are measured.

\subsubsection{Predicate Reliability Score}

The Predicate Reliability Score measures how consistently a predicate is preserved under a given degradation type and severity level. For binary or categorical predicates, reliability is defined as the agreement between clean and degraded predicate estimates, weighted by the confidence of the degraded estimate:
\begin{equation}
\mathrm{PRS}_{k}^{d,s}
=
\frac{1}{T}
\sum_{t=1}^{T}
c^{d,s}_{k,t}
\,
\mathbb{I}
\left(
\hat{y}^{d,s}_{k,t}
=
\hat{y}^{0}_{k,t}
\right),
\end{equation}
where $\mathbb{I}(\cdot)$ is the indicator function. A high value of $\mathrm{PRS}_{k}^{d,s}$ indicates that predicate $p_k$ remains stable and confident under degradation condition $(d,s)$.

For continuous predicates, such as distance-based proximity scores, agreement can be replaced by a normalized similarity function:
\begin{equation}
\mathrm{PRS}_{k}^{d,s}
=
\frac{1}{T}
\sum_{t=1}^{T}
c^{d,s}_{k,t}
\left(
1-
\frac{
\left|
\hat{y}^{d,s}_{k,t}-\hat{y}^{0}_{k,t}
\right|
}{
\alpha_k
}
\right)_{+},
\end{equation}
where $\alpha_k$ is a predicate-specific normalization constant and $(x)_{+}=\max(x,0)$. In both cases, higher $\mathrm{PRS}_{k}^{d,s}$ values indicate higher predicate reliability.

\subsubsection{Predicate Degradation Sensitivity}

The Predicate Degradation Sensitivity measures how strongly a predicate deteriorates as degradation severity increases. For a fixed degradation type $d$, let $s_{\min}$ and $s_{\max}$ denote the lowest and highest severity levels. The sensitivity of predicate $p_k$ is defined as
\begin{equation}
\mathrm{PDS}_{k}^{d}
=
\frac{
\mathrm{PRS}_{k}^{d,s_{\min}}
-
\mathrm{PRS}_{k}^{d,s_{\max}}
}{
s_{\max}-s_{\min}
}.
\end{equation}
A larger $\mathrm{PDS}_{k}^{d}$ indicates that predicate $p_k$ is more sensitive to degradation type $d$. For example, contact may have high sensitivity to occlusion or blur, while coarse around/proximity relations may remain more stable.

When multiple severity levels are available, a more general formulation estimates sensitivity as the average reliability drop between consecutive severity levels:
\begin{equation}
\mathrm{PDS}_{k}^{d}
=
\frac{1}{|\mathcal{S}|-1}
\sum_{r=1}^{|\mathcal{S}|-1}
\left(
\mathrm{PRS}_{k}^{d,s_r}
-
\mathrm{PRS}_{k}^{d,s_{r+1}}
\right),
\end{equation}
where $s_1 < s_2 < \cdots < s_{|\mathcal{S}|}$. Lower values indicate greater robustness to increasing degradation severity.

\subsubsection{Temporal Predicate Consistency}

Manipulation predicates should evolve coherently over time. A predicate sequence that changes abruptly due to missed detections, tracking noise, or frame dropping is less reliable than a temporally stable sequence. Temporal Predicate Consistency measures the smoothness of predicate estimates over time. For binary or categorical predicates, it is defined as
\begin{equation}
\mathrm{TPC\text{-}S}^{d,s}_{k}
=
1-
\frac{1}{T-1}
\sum_{t=2}^{T}
\mathbb{I}
\left(
\hat{y}^{d,s}_{k,t}
\neq
\hat{y}^{d,s}_{k,t-1}
\right).
\end{equation}
A higher value indicates fewer abrupt changes and greater temporal stability. However, because manipulation actions naturally contain meaningful relational transitions, this metric should be interpreted together with the clean predicate sequence. Therefore, we also define a transition-consistency version:
\begin{equation}
\mathrm{TPC}_{k}^{d,s}
=
1-
\frac{1}{T-1}
\sum_{t=2}^{T}
\left|
\mathbb{I}
\left(
\hat{y}^{d,s}_{k,t}
\neq
\hat{y}^{d,s}_{k,t-1}
\right)
-
\mathbb{I}
\left(
\hat{y}^{0}_{k,t}
\neq
\hat{y}^{0}_{k,t-1}
\right)
\right|.
\end{equation}
This formulation evaluates whether the degraded sequence preserves the same relational change pattern as the clean sequence, rather than merely rewarding a predicate for remaining constant.

\subsubsection{Confidence-Weighted Predicate Stability}

Since each predicate estimate is associated with a confidence score, the framework also measures the stability of confidence over time and degradation. Confidence-weighted stability is defined as
\begin{equation}
\mathrm{CWS}_{k}^{d,s}
=
\frac{1}{T}
\sum_{t=1}^{T}
c^{d,s}_{k,t}
\left(
1-
\left|
c^{d,s}_{k,t}
-
c^{0}_{k,t}
\right|
\right),
\end{equation}
where $c^{0}_{k,t}$ is the confidence of predicate $p_k$ in the clean video. This metric is high when a predicate remains both confidently estimated and close to its clean-condition confidence profile. It is useful for identifying predicates whose labels may remain unchanged but whose supporting evidence becomes weak under degradation.

\subsubsection{Downstream Impact Score}

Predicate reliability is important because predicate failures may affect downstream manipulation understanding. The Downstream Impact Score measures how much downstream performance changes when predicate $p_k$ is removed, corrupted, or replaced by its degraded estimate. Let $A_{\mathrm{all}}^{0}$ denote the downstream performance using all clean predicates, and let $A_{\setminus k}^{d,s}$ denote the performance when predicate $p_k$ is degraded, masked, or removed while the remaining predicates are kept available. The downstream impact of predicate $p_k$ is defined as
\begin{equation}
\mathrm{DIS}_{k}^{d,s}
=
A_{\mathrm{all}}^{0}
-
A_{\setminus k}^{d,s}.
\end{equation}
A larger $\mathrm{DIS}_{k}^{d,s}$ indicates that failures of predicate $p_k$ have a stronger effect on downstream manipulation understanding. The downstream measure $A$ may correspond to action-recognition accuracy, primitive prediction accuracy, event classification score, or another task-level performance measure.

In settings where the downstream model produces probabilities rather than only class labels, the impact can also be measured by the change in prediction confidence for the correct action class:
\begin{equation}
\mathrm{DIS}_{k}^{d,s}
=
\frac{1}{T}
\sum_{t=1}^{T}
\left|
\Pr(a_t \mid \Omega_t^{0})
-
\Pr(a_t \mid \Omega_{t\setminus k}^{d,s})
\right|,
\end{equation}
where $a_t$ is the target action or primitive label, $\Omega_t^{0}$ denotes the clean predicate state at time $t$, and $\Omega_{t\setminus k}^{d,s}$ denotes the degraded predicate state with predicate $p_k$ removed or corrupted. This formulation captures whether a predicate failure changes the confidence of downstream action interpretation.

\subsubsection{Aggregate Reliability Profile}

For each predicate $p_k$, the final reliability profile is defined as
\begin{equation}
\Phi_k^{d,s}
=
\left[
\mathrm{PRS}_{k}^{d,s},
\mathrm{PDS}_{k}^{d},
\mathrm{TPC}_{k}^{d,s},
\mathrm{CWS}_{k}^{d,s},
\mathrm{DIS}_{k}^{d,s}
\right].
\end{equation}
This profile summarizes whether a predicate is stable, degradation-sensitive, temporally consistent, confidence-preserving, and important for downstream understanding. Predicates with high $\mathrm{PRS}$ and high $\mathrm{TPC}$ are considered stable, predicates with high $\mathrm{PDS}$ are degradation-sensitive, and predicates with high $\mathrm{DIS}$ are high-impact predicates whose failures are especially important for manipulation understanding.

Table~\ref{tab:reliability_metrics} summarizes the proposed metrics and their interpretation.

\begin{table}[!b]
\centering
\caption{Proposed predicate reliability metrics.}
\label{tab:reliability_metrics}
\small
\begin{tabular}{p{2.2cm} p{5.0cm} p{4.2cm}}
\hline
\textbf{Metric} & \textbf{Purpose} & \textbf{Interpretation} \\
\hline
$\mathrm{PRS}$ 
& Measures predicate preservation under degradation 
& Higher is better \\

$\mathrm{PDS}$ 
& Measures sensitivity to increasing degradation severity 
& Lower is better \\

$\mathrm{TPC}$ 
& Measures temporal consistency of predicate transitions 
& Higher is better \\

$\mathrm{CWS}$ 
& Measures confidence preservation under degradation 
& Higher is better \\

$\mathrm{DIS}$ 
& Measures downstream effect of predicate failure 
& Higher means more critical predicate \\
\hline
\end{tabular}
\end{table}

\subsection{Framework Output and Downstream Use}
\label{subsec:framework_output}

The proposed framework outputs a structured reliability profile rather than only predicate labels. For a video condition $(d,s)$, the frame-level output is
\begin{equation}
\Omega_t^{d,s}
=
\left\{
\left(
p_k,
\hat{y}^{d,s}_{k,t},
c^{d,s}_{k,t}
\right)
\right\}_{k=1}^{K},
\end{equation}
where $p_k$ is the predicate, $\hat{y}^{d,s}_{k,t}$ is its estimated value, and $c^{d,s}_{k,t}$ is its confidence score at time $t$. Across the full video, these frame-level outputs form the temporal predicate sequence
\begin{equation}
\Omega^{d,s}
=
\left\{
\Omega_t^{d,s}
\right\}_{t=1}^{T}.
\end{equation}

At the predicate level, the framework outputs the reliability profile
\begin{equation}
\Phi_k^{d,s}
=
\left[
\mathrm{PRS}_{k}^{d,s},
\mathrm{PDS}_{k}^{d},
\mathrm{TPC}_{k}^{d,s},
\mathrm{CWS}_{k}^{d,s},
\mathrm{DIS}_{k}^{d,s}
\right],
\end{equation}
as defined in Section~\ref{subsec:predicate_reliability_metrics}. This profile indicates whether predicate $p_k$ is reliable under degradation, sensitive to increasing severity, temporally consistent, confidence-preserving, or important for downstream understanding. Aggregating over all predicates gives the video-level reliability summary
\begin{equation}
\Phi^{d,s}
=
\left\{
\Phi_k^{d,s}
\right\}_{k=1}^{K}.
\end{equation}

The reliability profile provides diagnostic information about the perception--predicate pipeline. Low contact reliability may indicate weak segmentation or boundary localization, whereas low reliability of dynamic predicates such as $\mathrm{GC}$, $\mathrm{MA}$, or $\mathrm{MT}$ may indicate tracking instability or temporal disruption. The profile therefore identifies which predicates require stronger perception modules, temporal smoothing, uncertainty modeling, or additional supervision.

The same output can be used by downstream manipulation-understanding systems. In action-recognition models, predicate values and confidence scores can be used as structured inputs together with visual features. In graph-based models, entities can be represented as nodes and predicates as typed edges, with confidence scores acting as edge weights. In neuro-symbolic or event-chain-based systems, reliable predicates can be converted into symbolic states, while uncertain predicates can be marked as $\mathrm{UNK}$ or handled using confidence-aware reasoning.

For downstream use, the predicate representation at time $t$ is written as
\begin{equation}
\Psi_t^{d,s}
=
\left[
\left(
\hat{y}^{d,s}_{1,t},c^{d,s}_{1,t}
\right),
\left(
\hat{y}^{d,s}_{2,t},c^{d,s}_{2,t}
\right),
\ldots,
\left(
\hat{y}^{d,s}_{K,t},c^{d,s}_{K,t}
\right)
\right].
\end{equation}
A downstream model $F(\cdot)$ may use this state to predict an action label, event state, or manipulation primitive:
\begin{equation}
\hat{a}^{d,s}_{t}
=
F\left(\Psi_t^{d,s}\right).
\end{equation}
The goal is not to prescribe the form of $F(\cdot)$, but to evaluate the quality of the predicate evidence supplied to it.

Thus, the framework enables degradation-aware reporting: instead of only stating that a model fails under blur, occlusion, or low resolution, it identifies which predicates fail, under which degradation type and severity, and how strongly those failures affect downstream interpretation.

\section{Experimental Protocol}
\label{sec:protocol}

This section presents the experimental protocol used to evaluate the proposed visual predicate reliability framework. The objective is to quantify how reliably manipulation predicates can be extracted from visual observations under controlled visual degradation, and to analyze how predicate failures affect downstream manipulation understanding. The protocol is designed to be predicate-centered, degradation-aware, and reproducible.

The evaluation follows the notation introduced in Section~\ref{sec:framework}. For each clean manipulation video $V^{0}$, degraded variants $V^{d,s}$ are generated for each degradation type $d \in \mathcal{D}$ and severity level $s \in \mathcal{S}$. Task-relevant entities are extracted from both clean and degraded videos, and visual predicates are computed over the same entity pairs or groups. The resulting predicate estimates $\hat{y}^{d,s}_{k,t}$ and confidence scores $c^{d,s}_{k,t}$ are compared against clean-reference or annotated predicate sequences to compute predicate reliability, degradation sensitivity, temporal consistency, confidence-weighted stability, and downstream impact.

The protocol consists of six main stages. First, manipulation videos are selected to cover diverse relational scenarios involving hand--object, object--container, object--surface, bimanual, and object--object interactions. Second, reference predicate sequences are constructed from clean videos, dataset annotations, or manual verification. Third, controlled visual degradations are applied at multiple severity levels. Fourth, entities and predicates are extracted using a fixed perception and predicate-extraction pipeline. Fifth, the proposed reliability metrics are computed for each predicate, degradation type, and severity level. Finally, the effect of predicate failures is evaluated in a downstream manipulation-understanding task.

The following subsections describe the datasets and action scenarios, predicate reference construction, visual degradation protocol, entity and predicate extraction implementation, evaluation metrics, downstream task, baselines, ablation studies, and experimental settings.

\subsection{Datasets and Action Scenarios}
\label{subsec:datasets}

The evaluation uses a combination of controlled and public manipulation datasets to analyze predicate reliability under both precisely defined and realistic visual conditions. The controlled set supports accurate predicate-reference construction and systematic degradation analysis, while public datasets test whether the proposed reliability profile generalizes to egocentric and bimanual manipulation videos.

We use four complementary data sources. The controlled manipulation set is recorded or curated under clean visual conditions and contains actions with clear relational transitions, such as approach, contact, grasp, transport, placement, insertion, removal, stacking, and release. VISOR/EPIC-KITCHENS provides egocentric video, object masks, and object-relation evidence, making it suitable for object--container, object--surface, and object--object predicates \cite{damen2022epic,darkhalil2022epic}. H2O provides first-person two-hand object interactions with hand pose, object pose, object categories, and interaction labels, supporting hand--object predicates such as contact, grasp, release, active-hand involvement, and coordinated motion \cite{kwon2021h2o}. ARCTIC provides dexterous bimanual manipulation with detailed 3D hand--object information, supporting fine-grained bimanual and contact-sensitive predicates \cite{fan2023arctic}. Broader activity datasets such as Ego4D, Ego-Exo4D, and Assembly101 are valuable for general video understanding \cite{grauman2022ego4d,grauman2024egoexo4d,sener2022assembly101}, but are not used as primary sources because the present study requires predicate-level references for contact, spatial configuration, motion coupling, grasp, release, containment, and support.

The selected action scenarios are organized into five relational groups: hand--object interaction, object--container interaction, object--surface interaction, bimanual manipulation, and object--object interaction. These groups cover both coarse relational changes, such as proximity, above/below, around, and stable configuration, and fine-grained changes, such as contact, grasp, release, containment, support, and moving together. This organization is important because different predicate groups are expected to exhibit different degradation-sensitivity profiles.

Let $\mathcal{A}=\{a_1,a_2,\ldots,a_M\}$ denote the set of manipulation action scenarios included in the evaluation. Each scenario $a_m$ contains a set of clean videos
\begin{equation}
\mathcal{V}_{a_m}=\{V^{0}_{a_m,n}\}_{n=1}^{N_m},
\end{equation}
where $V^{0}_{a_m,n}$ denotes the $n$-th clean video of scenario $a_m$, and $N_m$ is the number of clean sequences for that scenario. For each clean video, degraded variants are generated according to the degradation protocol described in Section~\ref{subsec:degradation_protocol}. The complete evaluation set is therefore defined as
\begin{equation}
\mathcal{V}
=
\left\{
V^{d,s}_{a_m,n}
\mid
a_m \in \mathcal{A},
n=1,\ldots,N_m,
d \in \mathcal{D},
s \in \mathcal{S}
\right\}.
\end{equation}

For each video, the relevant entity set is selected from the vocabulary in Section~\ref{subsec:visual_entity_extraction}. Predicate extraction is performed only over semantically meaningful entity pairs or groups for the corresponding scenario, such as hand--object pairs for grasp and release, object--container pairs for inside, and object--surface or object--object pairs for support. Table~\ref{tab:dataset_sources} summarizes the role of each dataset, while Table~\ref{tab:dataset_scenarios} summarizes the scenario organization and representative examples.

\begin{table*}[!t]
\centering
\caption{Datasets used in the predicate-level reliability evaluation.}
\label{tab:dataset_sources}
\small
\begin{tabularx}{\textwidth}{p{2.6cm} p{3.0cm} X X}
\hline
\textbf{Dataset} & \textbf{Role in evaluation} & \textbf{Relevant annotations or evidence} & \textbf{Predicate groups evaluated} \\
\hline

Controlled manipulation set
& Reference construction and controlled degradation analysis
& Manual or verified entity tracks, action phases, predicate transitions
& Contact, static spatial, dynamic spatial, grasp, release, support, containment \\

VISOR / EPIC-KITCHENS
& Realistic egocentric validation
& Object masks, egocentric video, object-relation evidence
& Object--container, object--surface, object--object, spatial and contact-related predicates \\

H2O
& First-person hand--object validation
& 3D hand pose, 6D object pose, object categories, interaction labels
& Hand--object contact, grasp, release, active hand, moving together \\

ARCTIC
& Fine-grained bimanual validation
& Dexterous bimanual hand--object information, 3D hand/object structure
& Bimanual interaction, contact-sensitive predicates, coordinated motion, role-dependent grasp/release \\

\hline
\end{tabularx}
\end{table*}

\begin{table*}[t]
\centering
\caption{Organization of manipulation scenarios for predicate-level reliability evaluation. Each row includes a representative example of the corresponding scenario group.}
\label{tab:dataset_scenarios}
\small
\setlength{\tabcolsep}{3pt}
\renewcommand{\arraystretch}{1.15}
\begin{tabularx}{\textwidth}{
>{\raggedright\arraybackslash}p{2.35cm}
>{\centering\arraybackslash}p{2.25cm}
>{\raggedright\arraybackslash}p{3.0cm}
>{\raggedright\arraybackslash}X}
\hline
\textbf{Scenario group} & 
\textbf{Representative example} & 
\textbf{Example actions} & 
\textbf{Key predicates} \\
\hline

Hand--object interaction
&
\includegraphics[width=2.05cm]{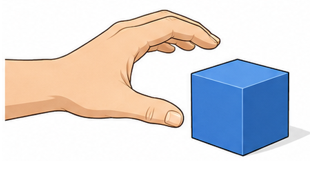}
&
approach, grasp, move, release
&
contact, non-contact, getting close, moving together, moving apart, grasp, release, active hand
\\

Object--container interaction
&
\includegraphics[width=2.05cm]{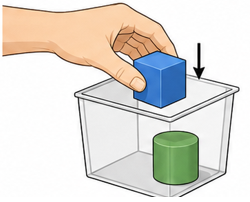}
&
insert, remove, pour, place into
&
inside, containment event, contact, support, above/below, stable configuration
\\

Object--surface interaction
&
\includegraphics[width=2.05cm]{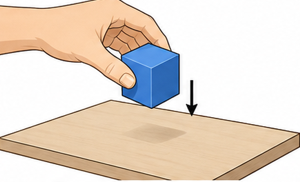}
&
place, lift, slide, stack
&
support/on-surface, above, contact, release, stable, moving together
\\

Bimanual manipulation
&
\includegraphics[width=2.05cm]{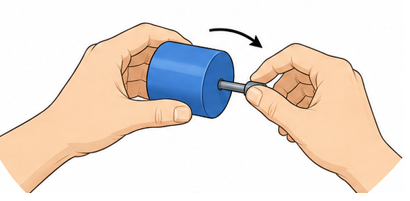}
&
transfer, hold-and-act, coordinated manipulation
&
left/right active hand, hand--object contact, coordinated motion, role-dependent grasp/release
\\

Object--object interaction
&
\includegraphics[width=2.05cm]{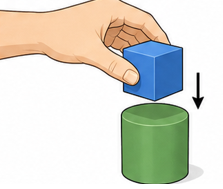}
&
stack, align, attach, separate
&
above/below, around, contact, support, moving apart, stable configuration
\\

\hline
\end{tabularx}
\end{table*}
\FloatBarrier

\subsection{Predicate Annotation and Reference Construction}
\label{subsec:predicate_reference}

Predicate reliability is evaluated by comparing predicate estimates under degradation against reference predicate sequences. For the controlled manipulation set, reference predicates are constructed from manual or semi-automatic annotations of entities, action phases, and relational transitions. Each sequence is annotated at the level of relevant entity pairs or groups, including hand--object, object--container, object--surface, and object--object relations. The reference includes contact relations, static spatial relations, dynamic spatial relations, and derived manipulation predicates such as grasp, release, support, containment, and active-hand involvement.

For public datasets, reference construction depends on the available annotations. In VISOR/EPIC-KITCHENS, object masks and object-level annotations are used to support spatial, containment, support, and object-relation predicates. In H2O, hand pose, object pose, object categories, and interaction labels are used to construct hand--object predicate references. In ARCTIC, 3D hand--object information is used to support fine-grained bimanual and contact-sensitive predicates. When direct predicate labels are not available, reference predicates are derived from the available annotations using the extraction rules defined in Section~\ref{subsec:predicate_vocabulary} and then manually verified for consistency.

For each clean video $V^{0}_{a_m,n}$, the reference predicate sequence is denoted as
\begin{equation}
Y^{0}_{a_m,n}
=
\left\{
y^{0}_{k,t}
\mid
p_k \in \mathcal{P},\; t=1,\ldots,T
\right\}.
\end{equation}
Here, $y^{0}_{k,t}$ denotes the reference value of predicate $p_k$ at time $t$. When high-quality manual annotation is available, $y^{0}_{k,t}$ is treated as ground truth. Otherwise, the clean-video estimate $\hat{y}^{0}_{k,t}$ is used as a clean reference, provided that its confidence score exceeds a predefined threshold $\theta_{\mathrm{ref}}$. Predicate instances with insufficient evidence are assigned the state $\mathrm{UNK}$ and are excluded from metrics that require a definite reference label.

Annotation quality is controlled in three ways. First, predicate definitions are fixed before annotation using the vocabulary and extraction rules in Section~\ref{subsec:predicate_vocabulary}. Second, ambiguous cases are conservatively assigned to $\mathrm{UNK}$ rather than forced into definite predicate states. Third, a subset of reference annotations is rechecked through repeated verification, and disagreements are resolved by inspecting the corresponding entity masks, tracks, and action phases. Annotation consistency can be assessed on a randomly selected subset of annotated sequences using percent agreement and Cohen's $\kappa$ for contact, static spatial, dynamic spatial, and derived predicates.

To reduce annotation ambiguity further, each predicate is evaluated only for semantically meaningful entity pairs or groups. For example, grasp and release are evaluated for hand--object pairs, inside is evaluated for object--container pairs, and support/on-surface is evaluated for object--surface or object--object pairs. This avoids penalizing the framework for predicates that are not meaningful in a given scenario.

\subsection{Visual Degradation Protocol}
\label{subsec:degradation_protocol}

To evaluate predicate reliability under realistic visual degradation, each clean video $V^{0}$ is transformed into degraded variants $V^{d,s}$ using a controlled degradation protocol. The degradation set $\mathcal{D}$ includes blur, occlusion, illumination change, low resolution, temporal frame dropping, and detection noise. Each degradation type is applied at five ordered severity levels,
\begin{equation}
\mathcal{S}=\{1,2,3,4,5\},
\end{equation}
where larger values correspond to stronger degradation. The clean condition is denoted by $s=0$ and is used as the reference condition. The same degradation parameters are applied consistently across videos within each severity level to ensure reproducibility.

Blur is implemented using Gaussian blur and motion blur to simulate defocus and camera or object motion. Occlusion is simulated by masking parts of hands, objects, containers, or surfaces, with the occluded area increasing across severity levels. Illumination degradation includes low-light transformation, overexposure, and contrast reduction. Low resolution is generated by downsampling frames and resizing them back to the original resolution. Temporal frame dropping removes or repeats frames to simulate reduced frame rate, packet loss, or temporal discontinuity. Detection noise is introduced at the perception-output level by perturbing bounding boxes, dropping detections, adding false detections, or inducing track interruptions.

For each degradation type, the severity parameter is chosen to produce a monotonic reduction in visual evidence while preserving the underlying action label. This is important because the goal is to measure degradation-induced predicate failure, not to create videos in which the manipulation action becomes semantically unrecognizable. The degradation parameters are summarized in Table~\ref{tab:degradation_protocol}. These values can be adjusted for dataset resolution, but the same severity schedule is used for all videos within a dataset.

\begin{table*}[t]
\centering
\caption{Visual degradation protocol used for predicate-level reliability evaluation. Severity increases from $s=1$ to $s=5$, while $s=0$ denotes the clean reference condition.}
\label{tab:degradation_protocol}
\footnotesize
\setlength{\tabcolsep}{3pt}
\renewcommand{\arraystretch}{1.12}
\begin{tabularx}{\textwidth}{
>{\raggedright\arraybackslash}p{2.2cm}
>{\raggedright\arraybackslash}p{3.6cm}
>{\raggedright\arraybackslash}p{3.4cm}
>{\raggedright\arraybackslash}X}
\hline
\textbf{Degradation} & 
\textbf{Severity parameter} & 
\textbf{Implementation} & 
\textbf{Predicates most affected} \\
\hline

Blur
& Kernel size or motion-blur length increases from mild to severe
& Gaussian blur and linear motion blur
& contact, grasp, release, boundary-based support \\

Occlusion
& Occluded area ratio increases across severity levels
& Partial masking of hands, objects, containers, or surfaces
& contact, active hand, containment, release \\

Illumination change
& Brightness and contrast factors vary from moderate to severe
& Low illumination, overexposure, contrast reduction
& segmentation-based and detection-based predicates \\

Low resolution
& Downsampling factor increases across severity levels
& Downsample frames and resize to original resolution
& contact, grasp, containment, fine spatial relations \\

Frame dropping
& Frame removal or repetition rate increases across severity levels
& Removed or repeated frames simulate temporal discontinuity
& dynamic predicates, release, moving together/apart \\

Detection noise
& Box perturbation, missed detections, false detections, and track breaks increase
& Perturb detections and tracks at the perception-output level
& all predicates, especially dynamic and derived predicates \\

\hline
\end{tabularx}
\end{table*}

\subsection{Entity and Predicate Extraction Implementation}
\label{subsec:implementation}

The same entity and predicate extraction pipeline is applied to clean and degraded videos to ensure fair comparison. The perception front-end produces entity labels, bounding boxes, masks, track identities, confidence scores, visibility estimates, and geometric states as defined in Section~\ref{subsec:visual_entity_extraction}. In the implemented pipeline, open-vocabulary localization is performed using Grounding DINO, mask extraction using Segment Anything, and temporal association using a tracking module \cite{liu2023grounding,kirillov2023segment}. Hand-specific evidence is obtained from dataset annotations when available, or from hand detection and hand-pose estimation when such annotations are missing.

For datasets with existing annotations, such as VISOR/EPIC-KITCHENS, H2O, and ARCTIC, the available masks, poses, object states, interaction labels, or 3D hand--object information are used whenever they provide more reliable entity evidence than automatic detection. Otherwise, the automatic perception front-end is used to estimate missing entities. This hybrid strategy exploits dataset-specific annotations while maintaining a consistent predicate-extraction procedure across controlled and public videos.

For each frame, task-relevant entities are selected from $E_t^{d,s}$, and predicate extraction is performed over valid entity pairs or groups. Contact predicates are estimated from mask overlap, boundary distance, or depth proximity when available. Static spatial predicates use centroid positions, bounding-box geometry, mask containment, and surface proximity. Dynamic predicates use trajectory distance, velocity similarity, and relative motion over a short temporal window. Derived predicates such as grasp, release, support events, containment events, and active-hand involvement are inferred from combinations of base contact, static spatial, and dynamic relations.

All predicate thresholds are selected on clean validation sequences and fixed for degraded conditions. This prevents degradation-specific tuning and ensures that changes in predicate reliability reflect changes in visual evidence rather than parameter adaptation. For each predicate estimate $\hat{y}^{d,s}_{k,t}$, the confidence score $c^{d,s}_{k,t}$ is computed using the evidence-quality factors introduced in Section~\ref{subsec:predicate_confidence}. Thus, the reported reliability scores reflect the behavior of the complete perception--predicate pipeline under degradation.

\subsection{Evaluation Metrics}
\label{subsec:evaluation_metrics}

The evaluation uses the reliability metrics defined in Section~\ref{subsec:predicate_reliability_metrics}. For each predicate $p_k$, degradation type $d$, and severity level $s$, we compute the Predicate Reliability Score $\mathrm{PRS}^{d,s}_{k}$, Predicate Degradation Sensitivity $\mathrm{PDS}^{d}_{k}$, Temporal Predicate Consistency $\mathrm{TPC}^{d,s}_{k}$, Confidence-Weighted Predicate Stability $\mathrm{CWS}^{d,s}_{k}$, and Downstream Impact Score $\mathrm{DIS}^{d,s}_{k}$. These metrics are reported at the levels of individual predicates, predicate groups, degradation types, and scenario groups.

For categorical predicates, reliability is measured by agreement with the reference predicate sequence; for continuous predicates, it is measured by normalized similarity to the reference value. Predicate instances labeled as $\mathrm{UNK}$ in the reference are excluded from agreement-based scores and analyzed separately as ambiguous or insufficient-evidence cases.

For each scenario group $g$, results are averaged over the corresponding video set $\mathcal{V}_g$. For example, the group-level Predicate Reliability Score is computed as
\begin{equation}
\overline{\mathrm{PRS}}^{d,s}_{k,g}
=
\frac{1}{|\mathcal{V}_g|}
\sum_{V \in \mathcal{V}_g}
\mathrm{PRS}^{d,s}_{k}(V),
\end{equation}
where $\mathcal{V}_g$ denotes the set of videos in scenario group $g$. The same aggregation is used for $\mathrm{TPC}$, $\mathrm{CWS}$, and $\mathrm{DIS}$. For $\mathrm{PDS}$, aggregation is performed after computing degradation sensitivity over severity levels for each video.

In addition to mean scores, variability is reported using standard deviation or confidence intervals where appropriate. Reliability curves and predicate-by-degradation heatmaps are used to show degradation trends and identify stable, sensitive, and high-impact predicates.

\subsection{Downstream Manipulation Understanding Task}
\label{subsec:downstream_task}

To evaluate whether predicate failures affect higher-level interpretation, the framework is tested in downstream manipulation-understanding tasks. The downstream evaluation is defined according to the available annotations in each dataset. In the controlled manipulation set, the task is manipulation primitive or event-state recognition, such as approach, grasp, transport, place, insert, remove, and release. In H2O and ARCTIC, the task is hand--object interaction or action recognition using the dataset-provided interaction labels. For VISOR/EPIC-KITCHENS, the downstream analysis focuses on object-centric action or relation-state interpretation when the required annotations are available.

The input to the downstream model is the confidence-weighted predicate state $\Psi_t^{d,s}$ defined in Section~\ref{subsec:framework_output}, optionally combined with visual features. The downstream model is denoted as
\begin{equation}
\hat{a}^{d,s}_{t}=F(\Psi_t^{d,s}),
\end{equation}
where $\hat{a}^{d,s}_{t}$ is the predicted action, manipulation primitive, or event state at time $t$. The purpose of this task is not to introduce a new recognition architecture, but to quantify how the quality of predicate evidence affects manipulation understanding. Therefore, the same downstream model $F(\cdot)$ is evaluated under clean and degraded predicate inputs.

Downstream impact is analyzed under three conditions. The first condition uses clean predicate inputs and provides the reference downstream performance. The second condition uses degraded predicate inputs, showing how visual degradation affects the downstream interpretation through predicate errors. The third condition removes, masks, or corrupts selected predicates to estimate their individual contribution to downstream performance. This allows the framework to identify high-impact predicates whose failure causes large changes in action interpretation.

The downstream performance measure $A$ is selected according to the task. For classification tasks, $A$ is reported using accuracy and macro-F1 score. For event-state or primitive recognition, frame-level or segment-level recognition accuracy is used. When the downstream model outputs class probabilities, the change in prediction confidence for the correct class is also reported. These measures are used to compute the Downstream Impact Score $\mathrm{DIS}^{d,s}_{k}$ defined in Section~\ref{subsec:predicate_reliability_metrics}.

\subsection{Baselines and Ablation Studies}
\label{subsec:baselines_ablations}

The experimental protocol includes baselines and ablations to isolate the contribution of predicate-level reliability analysis. The first baseline is an action-level robustness evaluation, where degradation effects are measured only through downstream task performance. This baseline tests whether predicate-level analysis provides diagnostic information beyond final accuracy or macro-F1 score. The second baseline uses unweighted predicate agreement, where predicate correctness is measured without confidence scores. The third baseline is frame-level predicate evaluation, where only per-frame predicate agreement is reported and temporal transition structure is ignored.

Ablation studies are conducted by removing or modifying key components of the proposed framework. The confidence ablation removes $c^{d,s}_{k,t}$ from reliability computation and evaluates whether confidence weighting improves diagnostic value. The temporal-consistency ablation removes $\mathrm{TPC}^{d,s}_{k}$ from the reliability profile while retaining the remaining metrics, testing whether transition-aware analysis contributes information beyond frame-level agreement. The degradation-sensitivity ablation evaluates predicate reliability at a single degradation level rather than across ordered severity levels, testing whether multi-severity analysis is necessary. The downstream-impact ablation removes predicate-level perturbation or masking and reports only global downstream performance, testing whether $\mathrm{DIS}^{d,s}_{k}$ identifies high-impact predicates that are not visible from aggregate action performance alone.

Table~\ref{tab:baselines_ablations} summarizes the baselines and ablations.

\begin{table}[t]
\centering
\caption{Baselines and ablation studies used in the evaluation.}
\label{tab:baselines_ablations}
\small
\begin{tabular}{p{3.2cm} p{7.0cm}}
\hline
\textbf{Method} & \textbf{Purpose} \\
\hline

Action-level robustness 
& Measures only downstream performance under degradation \\

Unweighted predicate agreement 
& Evaluates predicate correctness without confidence weighting \\

Frame-level predicate evaluation 
& Reports only per-frame predicate agreement and ignores temporal transitions \\

Without confidence 
& Removes $c^{d,s}_{k,t}$ from reliability computation \\

Without temporal consistency 
& Removes $\mathrm{TPC}^{d,s}_{k}$ from the full reliability profile \\

Without degradation sensitivity 
& Evaluates reliability at a single degradation level rather than across severity levels \\

Without downstream impact 
& Removes predicate perturbation/masking and reports only global downstream performance \\

\hline
\end{tabular}
\end{table}
\subsection{Experimental Settings}
\label{subsec:experimental_settings}

All experiments use fixed dataset splits, degradation parameters, entity extraction, predicate rules, confidence estimation, and downstream model settings across clean and degraded conditions. Predicate thresholds and confidence weights are selected on clean validation sequences and kept fixed for all degradation types and severity levels; no degradation-specific tuning is performed.

For each dataset, videos are divided at the sequence level into training, validation, and test subsets to avoid overlap between clips from the same manipulation instance. Unless an official split is provided, we use a $70\%/15\%/15\%$ split for training, validation, and testing. The validation set is used only to select predicate thresholds, confidence weights, temporal-window lengths, and downstream hyperparameters. All reported predicate-reliability and downstream results are computed on the held-out test set.

The degradation protocol uses one clean reference condition and five degradation severity levels:
\begin{equation}
s \in \{0,1,2,3,4,5\},
\end{equation}
where $s=0$ denotes the clean condition and $s=1,\ldots,5$ denote increasing degradation severity. For each clean test video, degraded variants are generated for every degradation type $d \in \mathcal{D}$ and severity level $s \in \{1,\ldots,5\}$ using the same severity schedule within each dataset.

Predicate thresholds are selected by maximizing agreement with reference labels on the clean validation set. For threshold-based predicates, such as contact, inside, support, and moving together, candidate thresholds are evaluated on a validation grid and the best-performing values are fixed for all test conditions. Confidence weights are selected on the same validation set and are not adjusted after degradation is applied.

Unless otherwise specified, dynamic predicates are computed over a local temporal window of $w=5$ frames, centered at the current frame when possible. This window is used for trajectory similarity, relative-motion estimation, and temporal consistency computation. For frame-dropping experiments, temporal indices are preserved to allow comparison with the corresponding clean-reference sequence.

Results are reported as averages over videos, scenario groups, predicates, degradation types, and severity levels. Variability is reported using standard deviation or $95\%$ confidence intervals where appropriate. The final analysis includes per-predicate reliability, degradation sensitivity, temporal consistency, confidence--reliability trends, downstream impact, ablations, and qualitative failure examples.




\section{Results}
\label{sec:results}

This section reports the experimental results of the proposed predicate-level reliability analysis. The evaluation is organized around five main questions: how reliably visual predicates are preserved under degradation, which predicates are most sensitive to each degradation type, whether temporal predicate transitions remain consistent, whether confidence scores reflect actual predicate reliability, and how predicate failures affect downstream manipulation understanding. Following the protocol in Section~\ref{sec:protocol}, results are reported across the controlled manipulation set and the public validation datasets, over degradation types $d \in \mathcal{D}$ and severity levels $s \in \{0,1,\ldots,5\}$, where $s=0$ denotes the clean reference condition. We analyze reliability at the level of individual predicates, predicate groups, degradation types, severity levels, and scenario groups. The results are summarized using the metrics defined in Section~\ref{subsec:predicate_reliability_metrics}, namely $\mathrm{PRS}$, $\mathrm{PDS}$, $\mathrm{TPC}$, $\mathrm{CWS}$, and $\mathrm{DIS}$.

\subsection{Overall Predicate Reliability under Degradation}
\label{subsec:overall_reliability}

\begin{figure}[t]
\centering
\includegraphics[width=\linewidth]{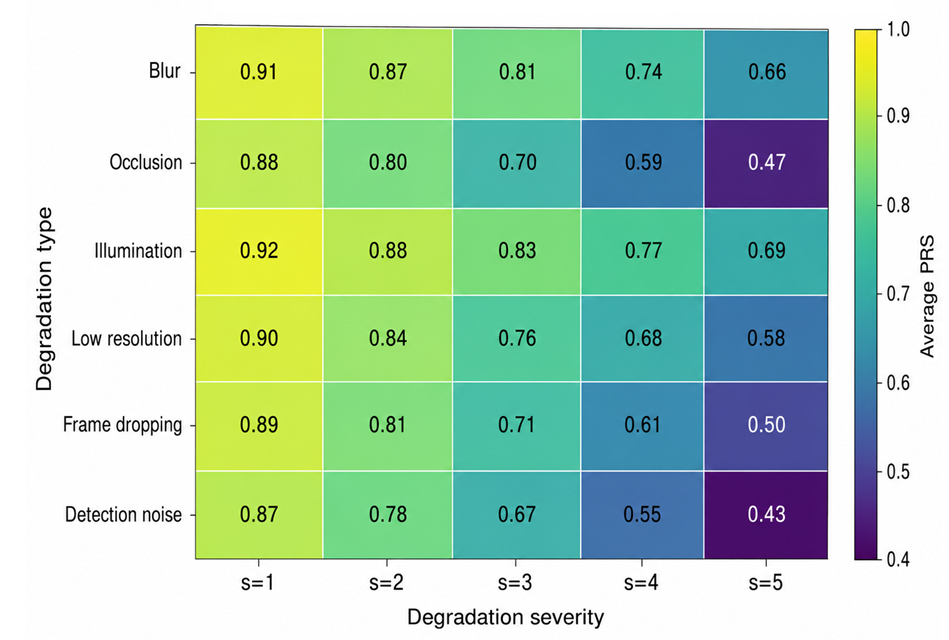}
\caption{Overall predicate reliability under visual degradation. The heatmap shows the average Predicate Reliability Score ($\mathrm{PRS}$) across all evaluated predicates, datasets, and scenario groups for each degradation type and severity level. Higher values indicate more reliable predicate extraction. The clean reference condition is not shown because it is shared across degradation types and achieves an average $\mathrm{PRS}$ of $0.94$.}
\label{fig:overall_reliability_heatmap}
\end{figure}

We first evaluate the overall reliability of visual predicates under different degradation types and severity levels. Figure~\ref{fig:overall_reliability_heatmap} provides a global view of the average Predicate Reliability Score ($\mathrm{PRS}$) across degraded conditions, while Table~\ref{tab:overall_prs_by_degradation} reports the corresponding numerical values. Since the clean reference condition is shared across degradation types, it is reported separately rather than repeated as a table column. This analysis identifies which degradation types cause the strongest overall reduction in predicate reliability before examining predicate-specific effects in the following subsections.

\begin{table}[!htbp]
\centering
\caption{Average Predicate Reliability Score ($\mathrm{PRS}$) across degradation types and severity levels. Higher values indicate more reliable predicate extraction. The clean reference condition achieves an average $\mathrm{PRS}$ of $0.94$ across all predicates and scenario groups.}
\label{tab:overall_prs_by_degradation}
\small
\setlength{\tabcolsep}{5pt}
\begin{tabular}{lccccc}
\hline
\textbf{Degradation} & \textbf{$s=1$} & \textbf{$s=2$} & \textbf{$s=3$} & \textbf{$s=4$} & \textbf{$s=5$} \\
\hline
Blur              & 0.91 & 0.87 & 0.81 & 0.74 & 0.66 \\
Occlusion         & 0.88 & 0.80 & 0.70 & 0.59 & 0.47 \\
Illumination      & 0.92 & 0.88 & 0.83 & 0.77 & 0.69 \\
Low resolution    & 0.90 & 0.84 & 0.76 & 0.68 & 0.58 \\
Frame dropping    & 0.89 & 0.81 & 0.71 & 0.61 & 0.50 \\
Detection noise   & 0.87 & 0.78 & 0.67 & 0.55 & 0.43 \\
\hline
\end{tabular}
\end{table}

The results show that predicate reliability decreases monotonically as degradation severity increases. Under mild degradation ($s=1$), all degradation types preserve relatively high reliability, with average $\mathrm{PRS}$ values above $0.87$. Under severe degradation ($s=5$), reliability drops substantially. Detection noise produces the lowest reliability ($0.43$), followed by occlusion ($0.47$) and frame dropping ($0.50$). In contrast, illumination change and moderate blur have a less severe effect on the average predicate profile. This suggests that predicate extraction is especially vulnerable when entity localization, visibility, and temporal continuity are disrupted.

To further analyze whether all predicate groups are affected similarly, Table~\ref{tab:overall_prs_by_group} reports the average $\mathrm{PRS}$ by predicate group, averaged across degradation types. In this table, the clean condition $s=0$ is retained because different predicate groups can have different clean-condition reliability.

\begin{table}[!htbp]
\centering
\caption{Average Predicate Reliability Score ($\mathrm{PRS}$) by predicate group, averaged across degradation types. Higher values indicate more reliable predicate extraction. Severity $s=0$ denotes the clean reference condition.}
\label{tab:overall_prs_by_group}
\small
\setlength{\tabcolsep}{5pt}
\begin{tabular}{lcccccc}
\hline
\textbf{Predicate group} & \textbf{$s=0$} & \textbf{$s=1$} & \textbf{$s=2$} & \textbf{$s=3$} & \textbf{$s=4$} & \textbf{$s=5$} \\
\hline
Contact relations          & 0.95 & 0.88 & 0.80 & 0.70 & 0.60 & 0.49 \\
Static spatial relations   & 0.96 & 0.93 & 0.89 & 0.84 & 0.78 & 0.70 \\
Dynamic spatial relations  & 0.93 & 0.87 & 0.78 & 0.68 & 0.57 & 0.46 \\
Derived predicates         & 0.92 & 0.84 & 0.74 & 0.62 & 0.50 & 0.39 \\
\hline
\end{tabular}
\end{table}

The predicate-group results show that coarse static spatial predicates remain relatively stable, whereas contact, dynamic, and derived manipulation predicates show larger degradation-induced drops. The strongest reliability is observed for static spatial relations, which remain comparatively robust even under severe degradation. This is expected because predicates such as above, below, around, and coarse support can often be estimated from approximate object geometry. In contrast, derived predicates such as grasp and release show the largest drop, reaching an average $\mathrm{PRS}$ of $0.39$ at severity level $s=5$. These predicates depend on multiple evidence sources, including contact, proximity, motion coupling, and temporal transition consistency; therefore, errors in any of these components can reduce the final predicate reliability.

Overall, these results support the central motivation of the paper: visual degradation does not affect all relational predicates uniformly. Instead, predicate reliability depends on the type of visual evidence required, the degradation mechanism that disrupts that evidence, and the temporal or compositional structure of the predicate.

\FloatBarrier

\subsection{Predicate-Specific Degradation Sensitivity}
\label{subsec:predicate_sensitivity}

We next analyze predicate-specific degradation sensitivity using the Predicate Degradation Sensitivity score ($\mathrm{PDS}$) defined in Section~\ref{subsec:predicate_reliability_metrics}. While Section~\ref{subsec:overall_reliability} showed the average reliability drop across all predicates, the present analysis identifies which individual predicates are most sensitive to each degradation type. This is important because two degradation conditions may produce similar average $\mathrm{PRS}$ values while affecting different relational cues.

\begin{figure}[t]
\centering
\includegraphics[width=\linewidth]{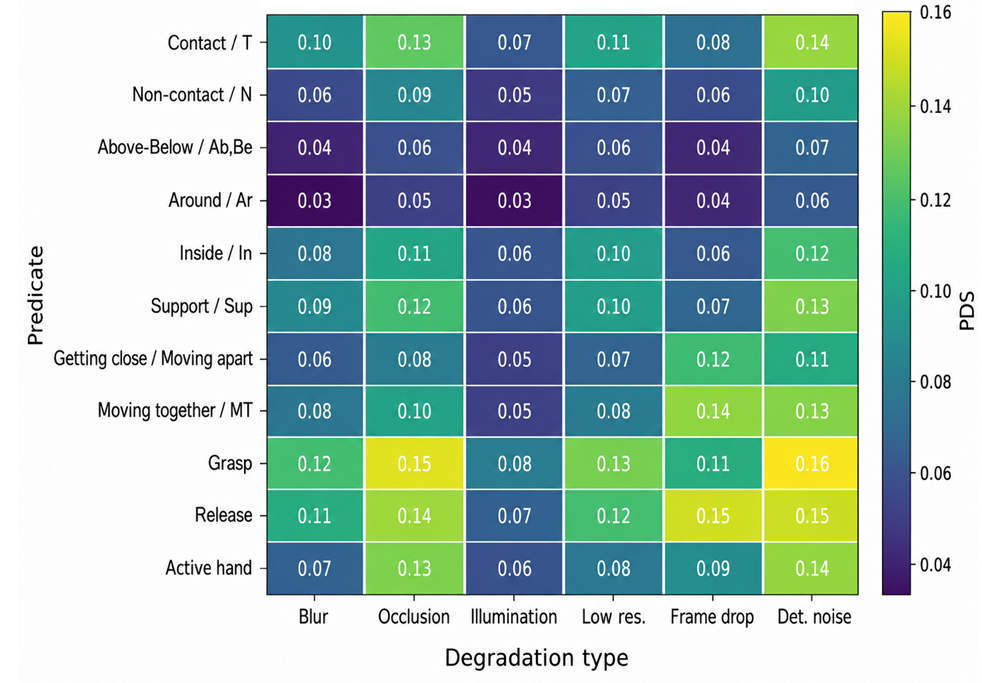}
\caption{Predicate-specific degradation sensitivity. The heatmap shows the Predicate Degradation Sensitivity score ($\mathrm{PDS}$) for representative predicates under each degradation type. Higher values indicate that the predicate reliability drops more rapidly as degradation severity increases.}
\label{fig:predicate_sensitivity_heatmap}
\end{figure}

Figure~\ref{fig:predicate_sensitivity_heatmap} provides a predicate-by-degradation view of $\mathrm{PDS}$, and Table~\ref{tab:predicate_pds} reports the corresponding numerical values. The clean condition $s=0$ is not included in this analysis because $\mathrm{PDS}$ measures the rate of reliability degradation across the ordered severity levels $s=1,\ldots,5$. Higher $\mathrm{PDS}$ indicates stronger sensitivity to increasing degradation severity, whereas lower values indicate greater robustness.

\begin{table*}[t]
\centering
\caption{Predicate Degradation Sensitivity ($\mathrm{PDS}$) for representative predicates under different degradation types. Higher values indicate stronger sensitivity to increasing degradation severity.}
\label{tab:predicate_pds}
\scriptsize
\setlength{\tabcolsep}{3pt}
\renewcommand{\arraystretch}{1.05}
\resizebox{\textwidth}{!}{%
\begin{tabular}{lccccccc}
\hline
\textbf{Predicate} &
\textbf{Blur} &
\textbf{Occ.} &
\textbf{Illum.} &
\textbf{Low res.} &
\textbf{Frame drop} &
\textbf{Det. noise} &
\textbf{Mean} \\
\hline
Contact / $\mathrm{T}$          & 0.10 & 0.13 & 0.07 & 0.11 & 0.08 & 0.14 & 0.105 \\
Non-contact / $\mathrm{N}$      & 0.06 & 0.09 & 0.05 & 0.07 & 0.06 & 0.10 & 0.072 \\
Above--below / $\mathrm{Ab,Be}$ & 0.04 & 0.06 & 0.04 & 0.06 & 0.04 & 0.07 & 0.052 \\
Around / $\mathrm{Ar}$          & 0.03 & 0.05 & 0.03 & 0.05 & 0.04 & 0.06 & 0.043 \\
Inside / $\mathrm{In}$          & 0.08 & 0.11 & 0.06 & 0.10 & 0.06 & 0.12 & 0.088 \\
Support / $\mathrm{Sup}$        & 0.09 & 0.12 & 0.06 & 0.10 & 0.07 & 0.13 & 0.095 \\
GC / MA                         & 0.06 & 0.08 & 0.05 & 0.07 & 0.12 & 0.11 & 0.082 \\
Moving together / $\mathrm{MT}$ & 0.08 & 0.10 & 0.05 & 0.08 & 0.14 & 0.13 & 0.097 \\
Grasp                           & 0.12 & 0.15 & 0.08 & 0.13 & 0.11 & 0.16 & 0.125 \\
Release                         & 0.11 & 0.14 & 0.07 & 0.12 & 0.15 & 0.15 & 0.123 \\
Active hand                     & 0.07 & 0.13 & 0.06 & 0.08 & 0.09 & 0.14 & 0.095 \\
\hline
\end{tabular}%
}
\end{table*}

The results show a clear separation between coarse spatial predicates and fine-grained manipulation predicates. Static spatial predicates such as $\mathrm{Ab}$, $\mathrm{Be}$, and $\mathrm{Ar}$ have the lowest mean sensitivity, indicating that approximate relative geometry can often be preserved even when the visual input is degraded. The around predicate is the most robust among the evaluated predicates, with a mean $\mathrm{PDS}$ of $0.043$, because it depends mainly on coarse spatial proximity rather than precise boundary evidence.

In contrast, contact-sensitive and event-like predicates show substantially higher sensitivity. Contact, support, inside, grasp, release, and active-hand involvement are strongly affected by occlusion and detection noise. This is consistent with the fact that these predicates depend on accurate entity localization, mask boundaries, visibility, and stable track identities. For example, grasp obtains the highest mean sensitivity score ($0.125$), followed by release ($0.123$). These predicates are derived from multiple base relations and temporal changes; therefore, errors in contact, motion coupling, or track continuity can propagate into the derived predicate estimate.

Different degradation types also affect different predicate families. Blur and low resolution mainly affect contact, support, grasp, and release because these predicates require fine boundary evidence. Occlusion strongly affects contact, inside, support, active-hand involvement, and grasp because it directly hides the entities or interaction regions needed for predicate estimation. Frame dropping has a stronger effect on dynamic predicates such as getting close, moving apart, moving together, and release because these predicates depend on temporal continuity. Detection noise produces the highest sensitivity for several predicates, including contact, grasp, release, and active hand, showing that localization errors and track instability can affect both low-level and derived predicate streams.

Overall, this analysis confirms that predicate degradation is structured rather than uniform. Coarse spatial predicates are comparatively robust, while contact-sensitive, temporally dependent, and derived manipulation predicates are more fragile. This supports the need for predicate-specific reliability analysis: a single average robustness score cannot reveal which relational cues are responsible for manipulation-understanding failures.

\subsection{Temporal Consistency Analysis}
\label{subsec:temporal_consistency_results}

We next evaluate whether predicate sequences preserve their temporal structure under degradation. This analysis uses the Temporal Predicate Consistency score ($\mathrm{TPC}$) defined in Section~\ref{subsec:predicate_reliability_metrics}, which measures whether the degraded predicate sequence preserves the transition pattern of the clean reference sequence. This is particularly important for manipulation understanding because actions such as grasp, release, insertion, removal, and transfer are defined not only by frame-level relations, but also by the timing of relational changes.

\begin{figure}[t]
\centering
\includegraphics[width=\linewidth]{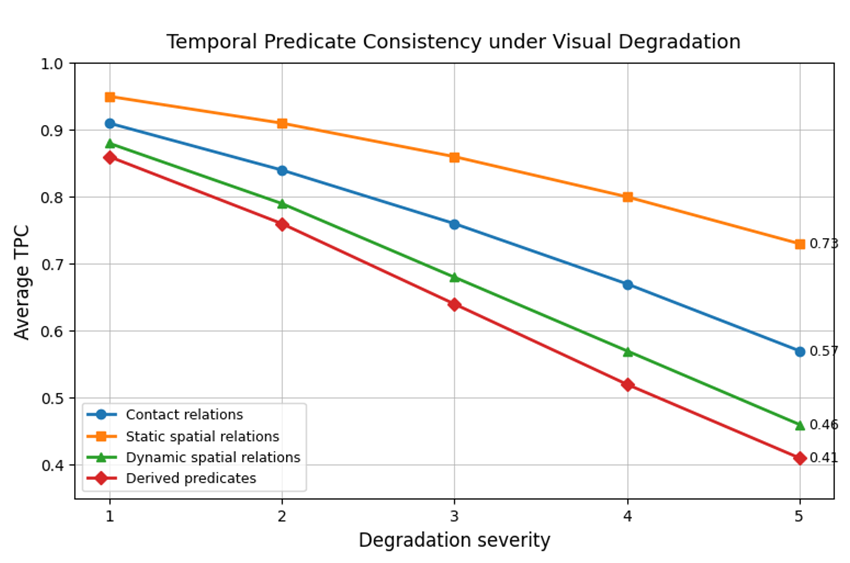}
\caption{Temporal predicate consistency under visual degradation. The curves show the average Temporal Predicate Consistency score ($\mathrm{TPC}$) for each predicate group across degradation severity levels. Higher values indicate better preservation of the clean-reference predicate transition pattern.}
\label{fig:temporal_consistency_curves}
\end{figure}

Figure~\ref{fig:temporal_consistency_curves} shows the average $\mathrm{TPC}$ over degradation severity for the four predicate groups. Table~\ref{tab:tpc_by_group} reports the corresponding numerical values. The results show that temporal consistency decreases as degradation severity increases, but the decrease is not uniform across predicate groups. Static spatial relations remain the most temporally stable, whereas dynamic spatial relations and derived predicates show the strongest degradation-induced temporal disruption.

\begin{table}[!htbp]
\centering
\caption{Average Temporal Predicate Consistency ($\mathrm{TPC}$) by predicate group, averaged across degradation types. Higher values indicate better preservation of clean-reference temporal transitions.}
\label{tab:tpc_by_group}
\small
\setlength{\tabcolsep}{5pt}
\begin{tabular}{lccccc}
\hline
\textbf{Predicate group} & \textbf{$s=1$} & \textbf{$s=2$} & \textbf{$s=3$} & \textbf{$s=4$} & \textbf{$s=5$} \\
\hline
Contact relations          & 0.91 & 0.84 & 0.76 & 0.67 & 0.57 \\
Static spatial relations   & 0.95 & 0.91 & 0.86 & 0.80 & 0.73 \\
Dynamic spatial relations  & 0.88 & 0.79 & 0.68 & 0.57 & 0.46 \\
Derived predicates         & 0.86 & 0.76 & 0.64 & 0.52 & 0.41 \\
\hline
\end{tabular}
\end{table}

The highest temporal consistency is observed for static spatial relations. This is expected because predicates such as above, below, around, and coarse support often change slowly and can remain temporally stable even when localization is moderately degraded. Contact relations show a larger drop because short missed detections, boundary ambiguity, or temporary occlusion can introduce spurious contact--non-contact transitions.

Dynamic spatial relations are more sensitive to degradation because they depend directly on trajectory continuity and relative motion. Their $\mathrm{TPC}$ decreases from $0.88$ at $s=1$ to $0.46$ at $s=5$, indicating that frame dropping, tracking instability, and detection noise can substantially distort motion-based predicate transitions. Derived predicates show the lowest temporal consistency at severe degradation, reaching $0.41$ at $s=5$. This is consistent with their compositional nature: grasp, release, containment events, and support events depend on both base predicate correctness and the timing of relational changes.

Overall, the temporal analysis shows that degradation affects not only whether a predicate is correct at individual frames, but also whether its transition structure remains meaningful over time. This supports the use of $\mathrm{TPC}$ as a complementary metric to $\mathrm{PRS}$ and $\mathrm{PDS}$, especially for dynamic and event-like manipulation predicates.

\subsection{Confidence Calibration and Reliability}
\label{subsec:confidence_calibration}

We next evaluate whether predicate confidence scores reflect actual predicate reliability. This analysis is important because the confidence score $c^{d,s}_{k,t}$ is intended to indicate the quality of the visual evidence supporting a predicate estimate, rather than merely providing an auxiliary output. A well-calibrated confidence signal should decrease when predicate estimates become unreliable under degradation.

\begin{figure}[t]
\centering
\includegraphics[width=\linewidth]{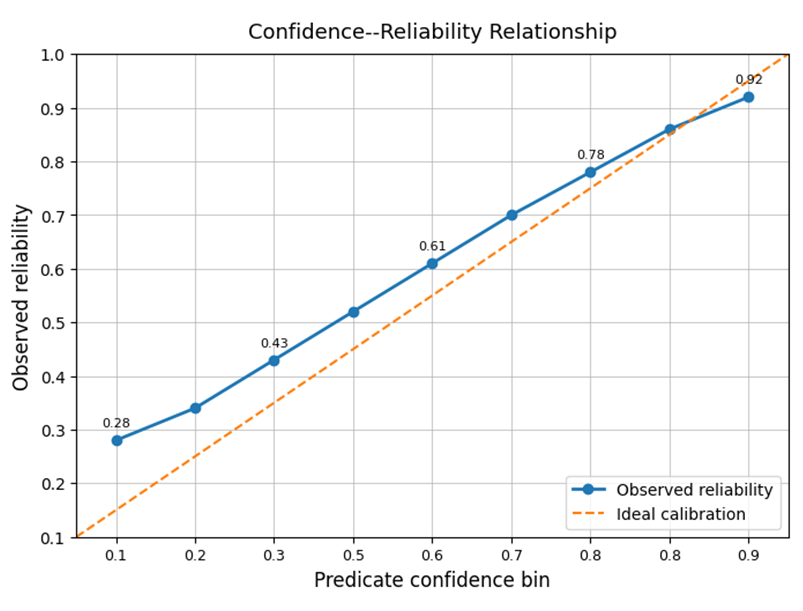}
\caption{Confidence--reliability relationship for visual predicates. Predicate estimates are grouped into confidence bins, and observed reliability is computed as the average agreement with the clean-reference predicate sequence within each bin. The monotonic trend indicates that confidence scores provide useful evidence about predicate reliability.}
\label{fig:confidence_calibration}
\end{figure}

Figure~\ref{fig:confidence_calibration} compares predicted confidence with observed predicate reliability across confidence bins. The observed reliability is computed as the average agreement with the clean-reference predicate sequence within each confidence interval. The results show a monotonic relationship between confidence and reliability: predicates assigned higher confidence are more likely to remain consistent with the clean reference.

\begin{table}[!htbp]
\centering
\caption{Confidence-weighted predicate stability ($\mathrm{CWS}$) by predicate group, averaged across degradation types. Higher values indicate that predicate estimates remain both confident and close to their clean-condition confidence profile.}
\label{tab:cws_by_group}
\small
\setlength{\tabcolsep}{5pt}
\begin{tabular}{lccccc}
\hline
\textbf{Predicate group} & \textbf{$s=1$} & \textbf{$s=2$} & \textbf{$s=3$} & \textbf{$s=4$} & \textbf{$s=5$} \\
\hline
Contact relations          & 0.90 & 0.82 & 0.73 & 0.63 & 0.52 \\
Static spatial relations   & 0.94 & 0.90 & 0.84 & 0.77 & 0.68 \\
Dynamic spatial relations  & 0.87 & 0.78 & 0.67 & 0.56 & 0.44 \\
Derived predicates         & 0.83 & 0.72 & 0.59 & 0.47 & 0.36 \\
\hline
\end{tabular}
\end{table}

Table~\ref{tab:cws_by_group} shows that confidence-weighted stability decreases with severity for all predicate groups. Static spatial relations preserve the highest $\mathrm{CWS}$, reaching $0.68$ even at $s=5$, which is consistent with their higher $\mathrm{PRS}$ and $\mathrm{TPC}$ values in Sections~\ref{subsec:overall_reliability} and~\ref{subsec:temporal_consistency_results}. In contrast, derived predicates show the lowest confidence-weighted stability, dropping to $0.36$ at $s=5$. This indicates that severe degradation not only changes their predicate values, but also weakens the evidence supporting them.

The confidence trends are also consistent with the predicate-specific sensitivity analysis in Section~\ref{subsec:predicate_sensitivity}. Predicates that depend on precise boundaries, stable tracks, or temporal transitions, such as contact, moving together, grasp, and release, lose confidence more rapidly under occlusion, frame dropping, and detection noise. Coarse spatial predicates remain better calibrated because their visual evidence is less dependent on fine boundary accuracy.

Overall, these results show that confidence scores provide meaningful reliability information. They can therefore be used not only for reporting predicate uncertainty, but also for downstream confidence-aware reasoning, where low-confidence predicates may be down-weighted or assigned to the $\mathrm{UNK}$ state instead of being treated as reliable symbolic evidence.

\subsection{Downstream Impact of Predicate Failures}
\label{subsec:downstream_impact_results}

We next evaluate whether predicate failures affect downstream manipulation understanding. Following Section~\ref{subsec:downstream_task}, the downstream model receives the confidence-weighted predicate state $\Psi^{d,s}_{t}$ and predicts the corresponding manipulation primitive or action state. We compare downstream performance under clean predicates, degraded predicates, and predicate-masked conditions. The analysis uses the Downstream Impact Score ($\mathrm{DIS}$) defined in Section~\ref{subsec:predicate_reliability_metrics}, where larger values indicate that failure of a predicate has a stronger effect on downstream interpretation.

\begin{figure}[t]
\centering
\includegraphics[width=\linewidth]{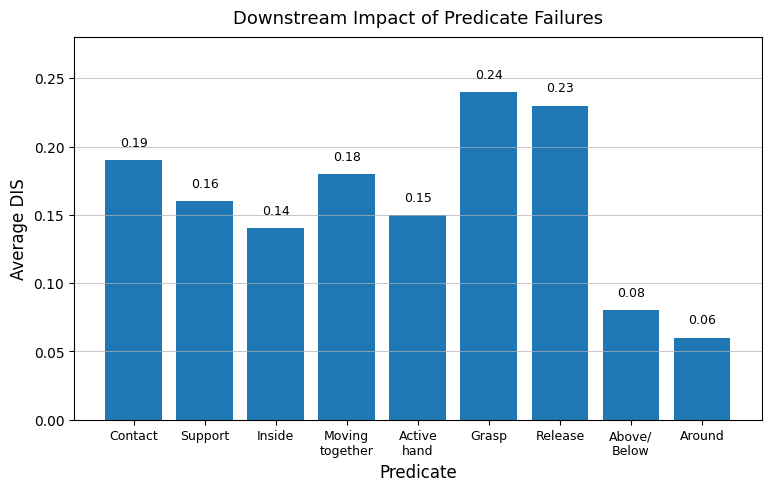}
\caption{Downstream impact of predicate failures. The bars show the average Downstream Impact Score ($\mathrm{DIS}$) for representative predicates. Higher values indicate that degrading or masking the predicate causes a larger drop in downstream manipulation-understanding performance.}
\label{fig:downstream_impact}
\end{figure}

Figure~\ref{fig:downstream_impact} shows the average $\mathrm{DIS}$ for representative predicates. The highest downstream impact is observed for grasp, release, contact, support, and moving together. These predicates are important because they directly distinguish manipulation primitives such as grasp, transport, place, insert, remove, and release. Thus, the most degradation-sensitive predicates identified in Section~\ref{subsec:predicate_sensitivity} are also among the most consequential for downstream understanding.

\begin{table}[!htbp]
\centering
\caption{Downstream manipulation-understanding performance under different predicate input conditions. Accuracy and macro-F1 are averaged across evaluated datasets and scenario groups.}
\label{tab:downstream_performance}
\small
\setlength{\tabcolsep}{4pt}
\begin{tabular}{lcc}
\hline
\textbf{Predicate input condition} & \textbf{Accuracy} & \textbf{Macro-F1} \\
\hline
Clean predicates                                      & 0.89 & 0.87 \\
Degraded predicates, $s=3$                             & 0.74 & 0.71 \\
Degraded predicates, $s=5$                             & 0.58 & 0.54 \\
Degraded predicates, $s=3$, without confidence weighting & 0.64 & 0.60 \\
Mask contact relations                                & 0.70 & 0.66 \\
Mask dynamic spatial relations                        & 0.68 & 0.63 \\
Mask derived predicates                               & 0.61 & 0.57 \\
\hline
\end{tabular}
\end{table}

Table~\ref{tab:downstream_performance} shows that downstream performance decreases as predicate reliability decreases. Clean predicate inputs achieve an average accuracy of $0.89$ and macro-F1 of $0.87$. Under moderate degradation ($s=3$), performance drops to $0.74$ accuracy and $0.71$ macro-F1. Under severe degradation ($s=5$), performance further decreases to $0.58$ accuracy and $0.54$ macro-F1. This confirms that degradation-induced predicate failures propagate to higher-level manipulation understanding.

The predicate-masking results show that different predicate groups have different downstream importance. Masking derived predicates produces the largest drop, reducing accuracy to $0.61$. This is consistent with the previous results, where derived predicates showed the lowest reliability, temporal consistency, and confidence-weighted stability under severe degradation. Masking dynamic spatial relations also causes a strong drop, reducing accuracy to $0.68$, because many manipulation primitives depend on motion coupling and temporal change. Contact relations are similarly important, especially for recognizing grasp, release, insertion, and placement events.

Confidence weighting improves downstream robustness. Under moderate degradation, removing confidence weighting reduces accuracy from $0.74$ to $0.64$ and macro-F1 from $0.71$ to $0.60$. This indicates that confidence scores help the downstream model reduce the influence of unreliable predicate estimates, especially when visual evidence is weakened by occlusion, frame dropping, or detection noise.

Overall, these results show that predicate reliability is not only a diagnostic property of the perception pipeline, but also a determinant of downstream manipulation-understanding performance. High-impact predicates require particular attention in robust system design because their failure can change the interpreted action even when other predicates remain reliable.

\subsection{Cross-Dataset Generalization}
\label{subsec:cross_dataset_results}

We next examine whether the predicate-level reliability trends generalize across datasets. The evaluation includes the controlled manipulation set and the three public validation sources described in Section~\ref{subsec:datasets}: VISOR/EPIC-KITCHENS, H2O, and ARCTIC. These datasets provide different forms of evidence, including segmentation masks, hand--object interaction labels, pose information, and bimanual manipulation structure. This analysis therefore tests whether the proposed reliability profile is specific to one dataset or reflects more general properties of manipulation predicates.

Table~\ref{tab:cross_dataset_results} reports average $\mathrm{PRS}$, $\mathrm{TPC}$, and downstream accuracy under moderate degradation ($s=3$), averaged across degradation types. The results show that the same overall pattern is preserved across datasets: predicate reliability decreases under degradation, but the framework remains able to identify stable and fragile predicate groups across different annotation formats and visual conditions.

\begin{table}[!htbp]
\centering
\caption{Cross-dataset generalization under moderate degradation ($s=3$). Scores are averaged across degradation types and evaluated predicate groups. Higher $\mathrm{PRS}$, $\mathrm{TPC}$, and downstream accuracy indicate better robustness.}
\label{tab:cross_dataset_results}
\small
\setlength{\tabcolsep}{5pt}
\begin{tabular}{lccc}
\hline
\textbf{Dataset} & \textbf{Avg. PRS} & \textbf{Avg. TPC} & \textbf{Downstream acc.} \\
\hline
Controlled set           & 0.78 & 0.77 & 0.78 \\
VISOR / EPIC-KITCHENS    & 0.72 & 0.69 & 0.71 \\
H2O                      & 0.75 & 0.72 & 0.74 \\
ARCTIC                   & 0.76 & 0.74 & 0.75 \\
\hline
\end{tabular}
\end{table}

The controlled set obtains the highest average reliability because entity tracks, action phases, and predicate references are more precisely defined. Among the public datasets, ARCTIC and H2O achieve slightly higher reliability than VISOR/EPIC-KITCHENS, mainly because their hand--object structure and pose-related information provide stronger evidence for contact, grasp, release, and coordinated motion. VISOR/EPIC-KITCHENS is more challenging because egocentric scenes include stronger background clutter, object diversity, partial visibility, and segmentation ambiguity.

Despite these dataset-specific differences, the relative behavior of predicate groups remains consistent with Sections~\ref{subsec:overall_reliability}--\ref{subsec:temporal_consistency_results}. Static spatial predicates remain the most robust, whereas dynamic and derived predicates are more sensitive to degradation. The downstream results follow the same trend: datasets with higher predicate reliability also preserve higher manipulation-understanding accuracy.

Overall, the cross-dataset analysis indicates that the proposed framework is not tied to a single dataset or annotation format. Although absolute reliability values vary with dataset difficulty and annotation quality, the relationship between predicate degradation, temporal consistency, and downstream performance remains stable across controlled, egocentric, and bimanual manipulation settings.
\subsection{Ablation Studies}
\label{subsec:ablation_results}

We conduct ablation studies to assess the contribution of the main components of the proposed reliability framework. The analysis is performed under moderate degradation ($s=3$), where predicate failures are visible but the manipulation action remains interpretable. Table~\ref{tab:ablation_results} summarizes the results.

\begin{table*}[!htbp]
\centering
\caption{Ablation results under moderate degradation ($s=3$). Predicate scores are averaged across degradation types, datasets, and predicate groups. Higher values indicate better reliability or downstream performance. A dash indicates that the corresponding diagnostic quantity is not available in that ablation.}
\label{tab:ablation_results}
\scriptsize
\setlength{\tabcolsep}{3pt}
\renewcommand{\arraystretch}{1.08}
\resizebox{\textwidth}{!}{%
\begin{tabular}{lccccc}
\hline
\textbf{Method} & \textbf{Pred. score} & \textbf{TPC} & \textbf{CWS} & \textbf{Accuracy} & \textbf{Macro-F1} \\
\hline
Full framework                    & 0.75 & 0.74 & 0.71 & 0.74 & 0.71 \\
Action-level robustness only       & --   & --   & --   & 0.74 & 0.71 \\
Unweighted predicate agreement     & 0.78 & --   & --   & 0.64 & 0.60 \\
Frame-level predicate evaluation   & 0.75 & --   & 0.71 & 0.68 & 0.65 \\
Without confidence weighting       & 0.75 & 0.74 & --   & 0.64 & 0.60 \\
Without temporal consistency       & 0.75 & --   & 0.71 & 0.69 & 0.66 \\
Without degradation sensitivity    & 0.75 & 0.74 & 0.71 & 0.74 & 0.71 \\
Without downstream-impact analysis & 0.75 & 0.74 & 0.71 & 0.74 & 0.71 \\
\hline
\end{tabular}%
}
\end{table*}

The full framework provides the most complete diagnostic profile, combining predicate preservation, temporal transition consistency, confidence-weighted stability, degradation sensitivity, and downstream impact. The action-level robustness baseline obtains the same downstream performance because it evaluates the same degraded predicate inputs, but it does not explain which predicates caused the performance loss. This confirms that final accuracy alone is insufficient for diagnosing manipulation-understanding failures.

Removing confidence information substantially reduces downstream performance. Without confidence weighting, accuracy drops from $0.74$ to $0.64$ and macro-F1 from $0.71$ to $0.60$, consistent with the confidence analysis in Section~\ref{subsec:confidence_calibration}. This shows that confidence scores help suppress unreliable predicate estimates rather than treating all predicate values as equally trustworthy.

The frame-level and temporal ablations show that transition-aware analysis is also important. Frame-level evaluation preserves the average predicate score but ignores whether relational changes occur at the correct time. As a result, downstream accuracy decreases to $0.68$. Similarly, removing temporal consistency reduces accuracy to $0.69$, indicating that manipulation primitives such as grasp, transport, release, insertion, and removal require temporally coherent predicate streams.

Removing degradation sensitivity does not change downstream performance in this setting because $\mathrm{PDS}$ is an analysis metric rather than a direct model input. However, without $\mathrm{PDS}$ the framework cannot rank predicates by robustness across severity levels. Similarly, removing downstream-impact analysis does not change predicate reliability scores, but prevents identification of high-impact predicates such as grasp, release, contact, and moving together.

Overall, the ablation results show that the proposed components are complementary. Predicate agreement measures whether relations are preserved, temporal consistency measures whether their transitions remain meaningful, confidence weighting estimates the trustworthiness of visual evidence, degradation sensitivity ranks predicate fragility, and downstream impact identifies predicates whose failures matter most for manipulation understanding.

\subsection{Qualitative Failure Analysis}
\label{subsec:qualitative_results}

\begin{figure*}[t]
\centering
\includegraphics[width=0.95\textwidth]{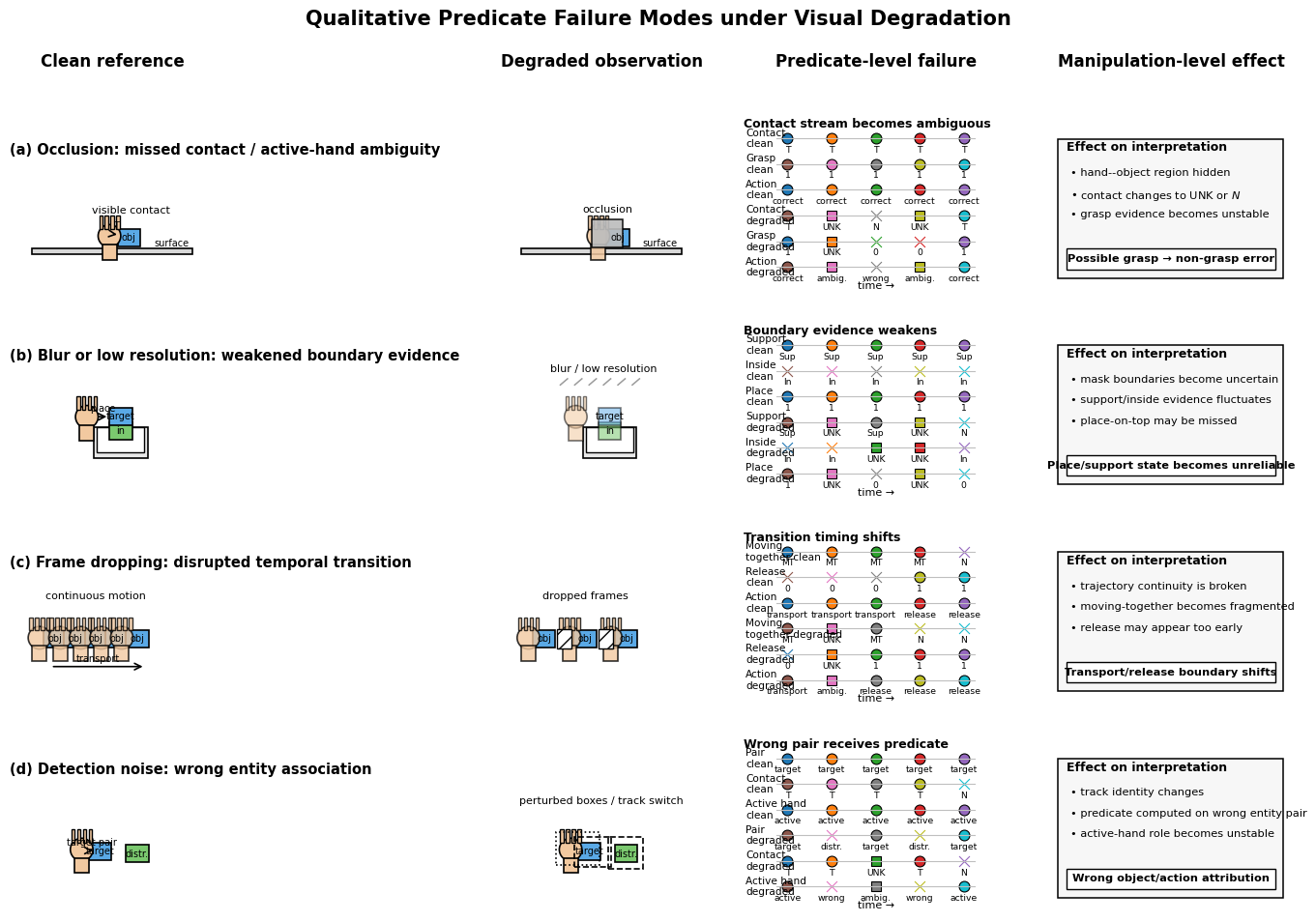}
\caption{Qualitative examples of predicate failures under visual degradation. Each example shows a clean reference frame or sequence, the degraded observation, the affected predicate stream, and the resulting manipulation-level interpretation. Typical failures include missed contact under occlusion, weakened grasp/support evidence under blur or low resolution, temporal disruption of moving-together and release under frame dropping, and incorrect predicate transitions under detection noise.}
\label{fig:qualitative_failures}
\end{figure*}

We finally provide qualitative examples to illustrate how predicate failures arise under visual degradation and how they affect manipulation interpretation. Figure~\ref{fig:qualitative_failures} shows representative cases selected from the evaluated scenario groups. The examples are chosen to cover contact-sensitive, spatial, dynamic, and derived predicates.

The first failure mode concerns contact-sensitive predicates. Under occlusion, the hand--object interaction region may be partially hidden, causing contact to be assigned as $\mathrm{UNK}$ or incorrectly changed to non-contact. This affects derived predicates such as grasp and release because they depend on reliable contact transitions. Such cases explain why contact and derived predicates show larger reliability drops in Sections~\ref{subsec:overall_reliability} and~\ref{subsec:predicate_sensitivity}.

The second failure mode involves degraded boundary evidence. Blur and low resolution weaken mask boundaries and reduce the precision of object geometry. As a result, support, inside, and grasp predicates may become unstable, especially when the manipulated object is small or partially overlapping with a container or supporting surface. These examples are consistent with the lower confidence-weighted stability of contact-sensitive and derived predicates reported in Section~\ref{subsec:confidence_calibration}.

The third failure mode is temporal disruption. Frame dropping and tracking instability can alter the apparent timing of relational changes. A release event may be delayed, missed, or fragmented into multiple short transitions, and moving-together may be confused with moving apart when trajectories become discontinuous. This explains the lower $\mathrm{TPC}$ values observed for dynamic spatial relations and derived predicates in Section~\ref{subsec:temporal_consistency_results}.

The fourth failure mode is entity-association error. Detection noise can perturb bounding boxes, introduce false detections, or break object tracks, which may cause predicates to be computed over the wrong entity pair. This is especially harmful in bimanual or cluttered egocentric scenes, where active-hand involvement and object--object relations depend on consistent identity assignment.

Overall, the qualitative analysis supports the quantitative findings. Coarse spatial predicates usually remain interpretable under moderate degradation, whereas contact-sensitive, dynamic, and derived predicates fail through boundary ambiguity, visibility loss, track discontinuity, or incorrect entity association. These examples also show why high-impact predicates such as contact, moving together, grasp, and release require confidence-aware and temporally consistent treatment before being used for downstream manipulation understanding.

\section{Discussion}
\label{sec:discussion}

The results support the central premise of this paper: robust manipulation understanding cannot be fully evaluated from final action accuracy alone. Manipulation interpretation depends on intermediate relational evidence, and this evidence can fail in structured ways under visual degradation. By analyzing predicate reliability directly, the proposed framework exposes failure modes that are hidden in conventional action-level robustness evaluation.

\subsection{Main Findings and Implications}
\label{subsec:main_findings_implications}

The main finding is that visual degradation affects manipulation predicates non-uniformly. Overall reliability decreases with degradation severity, but the failure pattern depends strongly on both the degradation type and the predicate type. Detection noise, occlusion, and frame dropping cause the strongest reliability losses, whereas illumination change and moderate blur are less damaging on average. This indicates that predicate extraction is particularly vulnerable when entity localization, visibility, and temporal continuity are disrupted.

The second finding is that predicate robustness is structurally organized. Static spatial predicates remain comparatively reliable because they can often be estimated from approximate geometry. In contrast, contact-sensitive, dynamic, and derived predicates are more fragile. Derived predicates such as grasp and release are especially vulnerable because they depend on multiple sources of evidence, including contact, proximity, motion coupling, and temporal transition structure. This confirms that predicate reliability cannot be inferred from a single global robustness score.

The third finding is that degradation affects temporal structure, not only frame-level correctness. Dynamic and derived predicates show strong temporal-consistency loss under severe degradation, demonstrating that frame dropping, tracking instability, and detection noise can alter the timing of relational changes. This is important because manipulation primitives such as grasp, transport, insertion, placement, and release are defined by relational transitions, not by isolated frame-level states.

The fourth finding is that confidence scores provide useful reliability information. Confidence-weighted stability follows the same trend as predicate reliability and temporal consistency, and removing confidence weighting reduces downstream performance under moderate degradation. This suggests that confidence should be treated as part of the predicate representation, rather than as an auxiliary score. Low-confidence predicates can be down-weighted, assigned to $\mathrm{UNK}$, or handled probabilistically before downstream reasoning.

Finally, the downstream analysis shows that predicate failures are not only diagnostic artifacts; they have direct consequences for manipulation understanding. Clean predicate inputs achieve substantially higher downstream performance than degraded predicate inputs, and masking derived, dynamic, or contact predicates produces large performance drops. Thus, the predicates that are most fragile under degradation are also among the most semantically important for interpreting manipulation actions.

These findings have practical implications. Robust manipulation systems should not treat all relational cues equally. High-impact predicates such as contact, moving together, grasp, support, and release require stronger perception modules, temporal verification, and uncertainty-aware reasoning. Conversely, more stable predicates such as coarse spatial relations can provide useful anchors when fine-grained evidence becomes unreliable. Predicate-level reliability therefore offers a principled way to prioritize model improvement, error diagnosis, and safety-aware manipulation interpretation.

\subsection{Relation to Graph-Based and Neuro-Symbolic Models}
\label{subsec:relation_reasoning_models}

The proposed framework is complementary to graph-based, event-chain-based, and neuro-symbolic manipulation models. Semantic Event Chains and enriched Semantic Event Chains represent manipulation actions through changes in spatial and contact relations over time, while graph-based models represent entities and their interactions as structured nodes and edges \cite{aksoy2011learning,ziaeetabar2017semantic,ziaeetabar2018recognition,ziaeetabar2020using,ziaeetabar2018prediction,ziaeetabar2024hierarchical}. More recent multimodal and neuro-symbolic extensions further use relational states, confidence-aware predicates, functional roles, or language-guided reasoning to support fine-grained manipulation interpretation \cite{ziaeetabar2025adaptive,ziaeetabar2026neurosymbolic}. These models demonstrate the value of relational representations, but they usually assume that the visual predicates supplied to the reasoning module are sufficiently reliable. The present work studies this assumption directly.

For graph-based models, predicate confidence can be interpreted as an edge reliability score. Instead of passing hard relation labels to a graph neural network, the system can use confidence-weighted predicate edges and reduce the influence of unreliable relations. For event-chain and symbolic models, uncertain predicates can be assigned to $\mathrm{UNK}$ rather than forced into incorrect symbolic states. For neuro-symbolic systems, the reliability profile $\Phi^{d,s}_{k}$ can provide an interface between visual perception and symbolic reasoning by indicating which predicates are stable, fragile, temporally inconsistent, or high-impact.

More broadly, graph-based visual reasoning has also been explored outside manipulation understanding, for example in multimodal medical image analysis where spatial and semantic graph relations can refine visual representations for dense prediction tasks~\cite{parisot2018disease,ma2022dgrunit,ziaeetabar2025efficientgformer}.

This perspective is important because the goal of the framework is not to replace recognition, graph reasoning, or neuro-symbolic architectures. Rather, it evaluates the quality of the relational evidence that such systems consume. In this sense, predicate reliability acts as a diagnostic layer between perception and reasoning. It makes explicit whether a downstream failure is likely caused by weak visual evidence, temporal disruption, confidence miscalibration, or the reasoning model itself.

This also clarifies the contribution of the paper. The work is not an incremental robustness benchmark for a particular detector, segmenter, or action classifier. It proposes a predicate-level evaluation paradigm for manipulation understanding, where the object of analysis is the trustworthiness of the relational evidence supporting action interpretation. This makes the framework applicable across perception backbones, datasets, and downstream reasoning architectures.

\subsection{Limitations and Future Work}
\label{subsec:limitations_future_work}

The study has several limitations. First, the measured reliability depends on the perception front-end used to extract entities, masks, tracks, and visual evidence. Although the framework is designed to be model-agnostic, different detectors, segmenters, or trackers may produce different absolute reliability values. The goal of the framework is therefore not to certify a universal predicate score, but to provide a systematic method for measuring predicate reliability under a fixed perception pipeline.

Second, some predicate references are derived from clean-video estimates or dataset annotations rather than exhaustive manual frame-level labeling. This is necessary for evaluating multiple datasets and degradation conditions, but it may introduce uncertainty for ambiguous predicates such as contact, support, containment, and release. The use of confidence thresholds and the $\mathrm{UNK}$ state reduces this problem, but future work should include larger manually verified predicate annotations and explicit inter-annotator agreement reporting.

Third, the degradation protocol evaluates degradation types separately. This improves interpretability because each failure source can be isolated, but real-world videos often contain mixed degradations, such as blur combined with occlusion, low illumination, and tracking instability. Future work should extend the protocol to combined and naturally occurring degradations, including compression artifacts, camera shake, viewpoint changes, and severe hand-object occlusion.

Fourth, the downstream task is used to measure the effect of predicate quality, not to introduce a new state-of-the-art recognition architecture. Therefore, the downstream results should be interpreted as evidence that predicate failures affect manipulation understanding, rather than as a benchmark of action-recognition performance.

Future work can extend the framework in three directions. The first direction is reliability-aware learning, where predicate confidence and degradation sensitivity are learned directly from data. The second is integration with graph neural networks, transformers, and neuro-symbolic reasoning modules, so that downstream models can adaptively use, ignore, or query uncertain predicates. The third is active reliability improvement, where high-impact unreliable predicates trigger targeted refinement, temporal smoothing, multimodal sensing, or human-in-the-loop verification. These extensions would move predicate reliability from a diagnostic tool toward an active component of robust and trustworthy manipulation understanding.

\section{Conclusion}
\label{sec:conclusion}

This paper presented a predicate-level reliability framework for robust manipulation understanding under visual degradation. Instead of evaluating manipulation systems only through final action accuracy, the proposed framework analyzes the reliability of intermediate visual predicates such as contact, support, containment, motion coupling, grasp, release, and active-hand involvement.

The framework defines a degradation-aware protocol, a structured predicate vocabulary, confidence-aware predicate estimation, and reliability metrics for predicate preservation, degradation sensitivity, temporal consistency, confidence-weighted stability, and downstream impact. Across controlled and public manipulation datasets, the results show that predicate failures are structured rather than uniform. Static spatial predicates remain comparatively robust, while contact-sensitive, dynamic, and derived predicates are more vulnerable to occlusion, detection noise, frame dropping, and boundary degradation.

The downstream analysis further shows that predicate reliability is not merely an internal diagnostic measure. Failures of high-impact predicates such as grasp, release, contact, support, and moving together can substantially reduce manipulation-understanding performance. Confidence weighting and temporal consistency analysis help identify and reduce the influence of unreliable predicate evidence.

Overall, the proposed framework provides a diagnostic layer for trustworthy manipulation understanding. It identifies which relational cues remain reliable, which fail under degradation, and which failures matter most for downstream interpretation. This predicate-level view can support more interpretable, robust, and uncertainty-aware manipulation systems, especially in egocentric video understanding, assistive robotics, human--robot collaboration, and graph-based or neuro-symbolic reasoning.

\section*{Data Availability}

The public datasets used in this study, including VISOR/EPIC-KITCHENS, H2O, and ARCTIC, are available through their respective project repositories and are cited in the manuscript. The controlled manipulation sequences, reference predicate annotations, generated visual-degradation configurations, and derived predicate-level evaluation results used in this study are available from the corresponding author upon request.

\section*{Funding Information}
This research received no specific grant from any funding agency in the public, commercial, or not-for-profit sectors.

\bibliography{sn-bibliography}
\end{document}